\documentclass{article}

\PassOptionsToPackage{numbers, compress}{natbib}
\usepackage[preprint]{neurips_2026}

\usepackage[utf8]{inputenc} 
\usepackage[T1]{fontenc}    

\usepackage{url}            
\usepackage{booktabs}       
\usepackage{amsfonts}       
\usepackage{nicefrac}       
\usepackage{microtype}      
\usepackage{xcolor}         
\usepackage{graphicx}
\usepackage{neurips_2026}
\usepackage{amsmath}
\usepackage{hyperref}       
\usepackage{courier}
\usepackage{changes}
\usepackage{wrapfig}

\usepackage{xspace}
\usepackage{multirow}
\usepackage[table]{xcolor}
\usepackage{algorithm}
\usepackage{algpseudocode}
\usepackage{float}
\usepackage{caption}
\newcommand{\good}[1]{\textcolor{green!60!black}{#1}}
\newcommand{\bad}[1]{\textcolor{red!70!black}{#1}}

\definecolor{LightBlue}{HTML}{D9E1F2}
\definecolor{LightGreen}{HTML}{E2F0D9}
\usepackage{booktabs}
\usepackage{multirow}
\usepackage[table]{xcolor}
\usepackage{array}
\usepackage{graphicx}
\usepackage{etoc}

\usepackage{multirow}
\usepackage{array}
\usepackage{booktabs}
\usepackage{makecell}
\usepackage{subcaption}
\usepackage{wrapfig}
\usepackage{tabularx}
\definecolor{gainGreen}{RGB}{0,150,0}

\newcolumntype{C}[1]{>{\centering\arraybackslash}m{#1}}
\newcolumntype{L}[1]{>{\raggedright\arraybackslash}p{#1}}
\newcolumntype{C}[1]{>{\centering\arraybackslash}p{#1}}

\usepackage{xspace}
\newcommand{\birdrl}{\textsc{Bird-RL}\xspace}
\newcommand{\birdtalon}{\textsc{Bird-Talon}\xspace}
\newcommand{\birdzeno}{\textsc{Bird-Zeno}\xspace}

\title{BIRD-RL: Scaling Agentic Reinforcement Learning over Stateful Data-Centric Environments}

\author{%
    Ge Qu$^{1}$, 
    Jinyang Li$^{1}$,
    Xiaohan Xu$^{1}$, 
    Xiaolong Li$^{1}$, \\
    \textbf{Bowen Qin$^{2}$}, 
    \textbf{Nan Huo$^{1}$}, 
    \textbf{Ziwei Tang$^{1}$}, 
    \textbf{Chenhao Ma$^{3}$}, 
    \textbf{Reynold Cheng$^{1}$} \\
    $^1$ The University of Hong Kong 
    $^2$ National University of Singapore \\
    $^3$ The Chinese University of Hong Kong, Shenzhen \\
    \texttt{quge@connect.hku.hk} \\
}

\begin{document}

\maketitle

\begin{abstract}
Modern data-centric applications increasingly require language agents capable of operating over live, stateful database environments, where successful problem resolution depends not merely on initial SQL generation, but on reasoning about evolving database states, interacting with execution feedback, and repairing state-mutating operations. While agentic reinforcement learning (RL) has emerged as a promising paradigm for training small language models (SLMs), existing frameworks are predominantly tool-centric and struggle with data-centric environments. Specifically, applying tool-centric frameworks to stateful tasks like SQL debugging introduces severe spatial inefficiencies since proper environment isolation typically requires a separate database container per rollout. Naively reusing databases without tracking evolving states across multi-turn interactions causes trajectory contamination and cross-trajectory interference, which severely destabilizes policy updates. 
To overcome these bottlenecks, we introduce \textbf{BIRD-RL}, a novel four-stage agentic RL curriculum that alternately enhances foundational agent reasoning and trajectory-adaptive on-policy exploration to circumvent the instability inherently associated with direct end-to-end learning. Crucially, to support this curriculum, BIRD-RL is driven by a \textbf{Trajectory-scoped Persistent Execution Infrastructure}, which enables safe database replica reuse while preserving trajectory isolation. This design improves spatial RL efficiency by \textbf{12.8x}, successfully enabling scalable agentic RL for complex data-centric tasks without needs of Kubernettes. Empirical evaluations on the BIRD-CRITIC-SQLite benchmark demonstrate that our 7B and 14B models achieve Success Rates of 44.40\% and 48.00\%, respectively, showing competitive performance against strong proprietary model-based agents such as Claude-Opus-4.6 Agent (48.20\%). Furthermore, BIRD-RL successfully trains unified SLMs capable of both advanced SQL debugging and generation, highlighting its broader potential for multi-objective optimization across heterogeneous, real-world data-centric tasks. For reproducibility, we release our code at \url{https://github.com/bird-bench/BIRD-RL}.
\end{abstract}

\section{Introduction}

Modern digital applications, from financial systems to web services, heavily rely on relational databases (RDBs) for robust data management~\cite{date2003, hellerstein2005readings, KandelPHH12, thusoosjsczalm10}. Structured Query Language (SQL), which functions as the definitive interface for these systems, is indispensable for data query and manipulation~\cite{chamberlin1974, armbrust2015}. Driven by the rapid development of large language models (LLMs)~\cite{singh2025openai, zeng2026glm}, a broad spectrum of LLM-based SQL tasks has emerged. Text-to-SQL, which translates natural language questions into executable SQL queries, has become a representative beneficiary of this progress~\cite{li2024bird, spider-2018, pourreza2023din, li2024codes}. While current models excel at direct SQL generation, effectively resolving SQL issues remains a significant challenge with the evolving user demands in realistic scenarios~\cite{ahadi2016students, Miedema2021, gathanilb20, wangknl20, haroonog24}. Such SQL issue resolution often requires considerable manual effort, given the complexity of query semantics, intricate underlying schemas, and heterogeneous database operations. As a result, automating this process offers substantial practical value by improving efficiency and reducing reliance on specialized expertise \cite{pavlo2017self, 10.14778/3675034.3675043}.
Automatic issue resolution inherently requires stronger agentic capabilities to conduct root cause location and repair through iterative interaction with the database environment~\cite{li2026swesql, yang2024swe, yang2023intercode}. Contemporary approaches to developing reliable debugging agents can be categorized into two primary paradigms: inference-time prompting~\cite{lei2025spider, yang2023intercode, yao2023react} and post-training-based agent learning~\cite{li2026swesql}. The former employs elaborate in-context orchestrations, in which LLMs iteratively refine intermediate reasoning decisions based on execution feedback. The latter paradigm strengthens small language models (SLMs) by internalizing these agentic capabilities directly into model parameters through supervised fine-tuning (SFT). Such agents are highly valuable in data-centric applications due to data privacy and security concerns~\cite{qu-etal-2025-share}. This practical need motivates the focus of our work.

Despite its effectiveness, SFT often fails to promote robust intermediate reasoning and generalizable repair behavior due to its heavy reliance on demonstrated correction trajectories \cite{chu2025sft, gudibande2024the}. Agentic reinforcement learning (RL), instead, has recently emerged as a promising paradigm for training interactive agents, enabling models to improve reasoning through autonomous exploration driven by environmental reward signals~\cite{zeng2026glm, bai2024digirl, zhang2026prorlagentrolloutasaservicerl}. However, agentic RL for SQL issue resolution presents unique, domain-specific challenges. On the one hand, SQL issue resolution requires a broader action space for root-cause analysis in declarative SQL. Agents often need to verify hypotheses through stateful database operations, including data manipulation (\texttt{INSERT}, \texttt{UPDATE}, \texttt{DELETE}) and schema modification (\texttt{ALTER TABLE})~\cite{date2003, huo2026birdinteract}, placing a strong demand on the underlying infrastructure to support persistent and isolated database interactions at scale. General tool-centric frameworks often manage environments via a container-per-rollout mechanism~\cite{verl, yang2024swe, liu2024agentbench}. However, applying this to data-intensive database environments incurs severe spatial inefficiencies and prohibitive scaling bottlenecks during massive trajectory generation. On the other hand, SQL debugging is a complicated task that simultaneously demands schema-grounded reasoning and dynamic environmental interaction. Direct end-to-end learning may overwhelm the policy within an exponentially large search space, resulting in suboptimal debugging effectiveness \cite{rl_survey}. 

\begin{figure*}
    \centering
    \includegraphics[width=\textwidth]{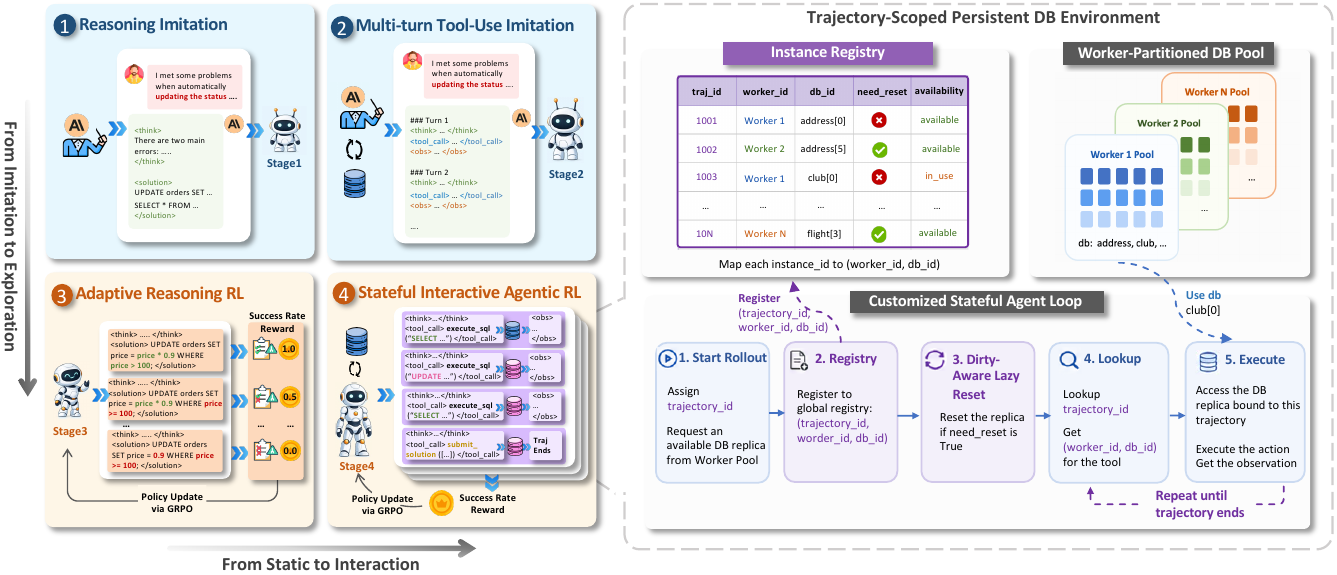}
    \caption{Overview of the \birdrl framework. \textbf{(Left)} The four-stage agentic RL curriculum transitioning from static reasoning imitation to interactive stateful exploration. \textbf{(Right)} The Trajectory-Scoped Persistent Execution Infrastructure, enabling scalable and isolated database replica reuse via a customized agent loop and worker-partitioned pool.}
    \label{fig:main_fig}
\vspace{-0.5 cm}
\end{figure*}

To bridge this gap, we introduce \birdrl, a novel agentic RL curriculum for SQL issue solving. It decomposes agent learning into four progressive stages: reasoning imitation (Section \ref{sec:stage1}), tool-use imitation (Section \ref{sec:stage2}), adaptive reasoning reinforcement (Section \ref{sec:stage3}), and interactive agentic reinforcement (Section \ref{sec:stage4}). Such a framework decouples an otherwise densely intertwined skill set into atomic, manageable learning objectives, circumventing the instability inherently associated with direct end-to-end learning. To realize scalable training for the final stage, we further introduce a \textbf{Trajectory-Scoped Persistent Execution Infrastructure} comprising a worker-partitioned database pool (Section \ref{sec:database_pool}) and a customized agent loop (Section \ref{sec:agent_loop}), enabling isolated stateful interactions and high-concurrency efficiency via the orchestrated reuse of database replicas.

Building on \birdrl, we release the \birdtalon model family, including 7B and 14B versions. Furthermore, to demonstrate that our curriculum generalizes beyond single-objective optimization, we introduce \birdzeno-7B, a unified multi-task model trained under a principled data-mixing ratio for heterogeneous SQL tasks targeting both debugging and query generation. As a resource contribution to the community, we also release \textsc{SIX-GYM-SQLite}, an automatically constructed training environment that extends \textsc{SIX-GYM} \cite{li2026swesql} to the SQLite dialect to facilitate future research for SQL agents.

Empirical evaluations demonstrate that \birdtalon-7B and 14B achieve Success Rates (SR) of \textbf{44.40\%} and \textbf{48.00\%} on \textsc{BIRD-CRITIC-SQLite}, respectively, performing competitively against strong proprietary model-based agents such as Claude-Opus-4.6 Agent (48.20\%). Our trajectory-scoped infrastructure improves RL spatial efficiency by \textbf{12.8×} under the evaluated SQLite workload, enabling stable and scalable agentic training. Besides, the unified \birdzeno model achieves an Execution Accuracy (EX) of \textbf{63.90\%} on the BIRD development set, delivering generation performance comparable to dedicated text-to-SQL models specifically optimized with RL while maintaining its debugging proficiency. These findings establish \birdrl as a scalable framework for data-centric agent training, with its trajectory-scoped infrastructure paving the way for agentic RL in heavy, stateful database environments, while underscoring the potential of multi-objective training for heterogeneous real-world SQL tasks.

\section{Preliminaries}
\subsection{Task Definition}
\paragraph{Text-to-SQL Generation.} 
Given a natural language question $\mathcal{Q}$, with its corresponding database schema $\mathcal{S}$, the goal of text-to-SQL is to generate the SQL query $\sigma_{pred}$ that could answer the question $\mathcal{Q}$ accurately by:
\begin{equation}
\sigma_{pred} = f_{\mathcal{\theta}}(\mathcal{Q},\mathcal{S}),
\label{eq1}
\end{equation}
where $f_{\mathcal{\theta}}\left(\cdot \right)$ refers to the mapping function with parameters $\theta$ applied by the generator.

\paragraph{SQL Issue Resolution.}
Compared with SQL generation from scratch, SQL issue resolution is a more realistic and challenging task. The task takes as input a user-provided SQL query $\sigma_{issue}$, a natural language problem description $\mathcal{P}$ specifying the issue and intended behavior, and the database schema $\mathcal{S}$. The goal is to produce a revised query $\sigma_{pred}$ that resolves the issue while preserving the user intent by:
\begin{equation}
\sigma_{pred} = f_{\mathcal{\theta}}(\mathcal{P},\mathcal{S},\sigma_{issue}),
\label{eq2}
\end{equation}
The target output $\sigma_{pred}$ is expected to satisfy the underlying user intent inferred from the input triplet.

\subsection{Tool-Centric Agent RL Loop}
\label{sec:tool-centric}

Given an SQL issue task $x=(\mathcal{P}, \mathcal{S}, \sigma_{\mathrm{issue}})$, a standard tool-centric agent loop \cite{verl} models interaction as a history-conditioned decision process. The agent maintains a textual history $h_t$, initialized as $h_0=x$, and at each step samples a structured tool invocation $a_t \in \mathcal{A}$ from policy $\pi_\theta$:
\begin{equation}
a_t \sim \pi_\theta(\cdot \mid h_t), \quad
o_t \sim f_k(\cdot \mid a_t, e_t), \quad
h_{t+1}=h_t \oplus (a_t,o_t),
\end{equation}
where $f_k$ denotes the selected tool, $o_t$ is the returned observation under environment state $e_t$, and $\oplus$ denotes history concatenation. The trajectory ends when the agent submits a final answer or reaches the maximum step budget.

This fundamentally tool-centric loop records tool feedback in the textual context while leaving the persistent environment external to the loop. For many agentic RL tasks \cite{verl, yang2024swe, liu2024agentbench}, isolation can be approximated by assigning each rollout an independent sandbox. In data-centric tasks such as SQL issue resolution, however, this \textit{container-per-rollout} design becomes prohibitively expensive, since each trajectory would require its own mutable replica of a potentially large database.

Reusing database replicas is therefore necessary for scalability, but remains unsafe under a purely tool-centric loop. Without explicit trajectory-to-environment binding, tool calls from the same rollout may fail to access the intended evolving database state, while reused replicas may carry residual mutations from previous rollouts. Such state leakage contaminates execution feedback and induces cross-trajectory interference, thereby destabilizing on-policy RL. These limitations motivate the Trajectory-scoped Persistent Execution Infrastructure introduced in Section~\ref{sec:infra}.

\section{\birdrl: Efficient Agentic RL Curriculum for SQL Agent Training}

Autonomous SQL issue resolution requires diagnostic reasoning, tool invocation, and feedback-guided refinement. Directly optimizing these capabilities end-to-end over a large interactive action space can lead to unstable or suboptimal policies \cite{rl_survey}. To address this, \birdrl adopts a four-stage curriculum, as shown in Figure~\ref{fig:main_fig} (left), which contains Reasoning Imitation Section \ref{sec:stage1}, Tool-Use Imitation (Section \ref{sec:stage2}), Reasoning Reinforcement (Section \ref{sec:stage3}), and Interactive Agentic Reinforcement (Section \ref{sec:stage4}). This progressive design decomposes agent learning from static reasoning to stateful interaction, while the final stage is supported by our trajectory-scoped infrastructure for scalable and contamination-free RL.


\subsection{Stage 1: Reasoning Imitation}
\label{sec:stage1}
\paragraph{Training Objective.} As the foundation of the curriculum, this initial stage aims to establish a strong reasoning prior for schema-grounded issue solving. Formulated as a single-turn sub-task, the target output for a given input task $x = (\mathcal{P}, \mathcal{S}, \sigma_{issue})$ is defined as $y = (t, \sigma)$, where $t$ denotes the reasoning thought, and $\sigma$ is the predicted query. The policy $\pi_\theta$ is trained via supervised fine-tuning (SFT). The structured diagnostic reasoning capability learned in Stage 1 effectively mitigates the naive exploration (e.g., \texttt{SELECT * FROM table}) in the ultimate interactive scenarios, resulting in more efficient execution and preventing unnecessary context expansion.

\paragraph{Training Data Construction.}
To reduce human annotation cost, we use \texttt{Claude-Sonnet-4.5} as the teacher model to construct reasoning-imitation data in the \textsc{SIX-GYM-SQLite} environment. Details of the environment construction and the protocol for preventing training-evaluation contamination are provided in Appendix~\ref{app:six_gym}. For each task $x=(\mathcal{P}, \mathcal{S}, \sigma_{\mathrm{issue}})$, the teacher generates a debugging thought $t$ with access to the corrected query $\sigma^*$, which serves as a reference for constructing the reasoning that diagnoses the SQL errors and motivates the target repair. We pair the generated thought with the corrected query as the supervision target $y=(t,\sigma^*)$. To ensure high-quality training data, we only include $(t,\sigma^*)$ pairs where the corrected SQL $\sigma^*$ can be regenerated from the debugging thought $t$ and the regenerated solution passes all test cases in $\mathcal{C}$. This produces \textbf{4,982} instances for the Reasoning Imitation stage.

\subsection{Stage 2: Multi-Turn Tool-Use Imitation}
\label{sec:stage2}

\paragraph{Training Objective.}
Building on the previous stage, Stage 2 performs trajectory fine-tuning with the SQL-Act scaffold~\cite{li2026swesql}, where executable SQL statements serve as fine-grained actions for database inspection, hypothesis testing, and root-cause localization. Each trajectory $\tau=\{(t_i,a_i,o_i)\}_{i=1}^{n}$ interleaves the agent thought $t_i$, SQL action $a_i$, and database observation $o_i$. We implement SQL actions with two tools: \texttt{execute\_sql} for exploration and testing, and \texttt{submit\_solution} for termination. These tools keep SQL as the core action while providing an explicit stopping signal. Tool descriptions are detailed in Figure \ref{fig:sql_debug_sysprompt}.

\paragraph{Data Construction.} Following the trajectory formulation defined above, the teacher model \texttt{Claude-Sonnet-4.5} is guided to generate a verified trajectory $\tilde{\tau}$ for each input task suite $x$. We further employ the f-plan boosting strategy \cite{li2026swesql} to augment successful candidate trajectories. We consider the query $\sigma$ submitted via \texttt{submit\_solution} as the predicted query of each trajectory and apply the identical verification protocol as Stage 1 to filter $(x,\tilde{\tau})$ pairs. We ultimately collected a training set of \textbf{4,502} trajectories. 

\subsection{Stage 3: Adaptive Reasoning Reinforcement}
\label{sec:stage3}

\paragraph{RL Training Objective.}
Although the first two stages provide reasoning and tool-use priors, they are learned from offline imitation and may induce rigid patterns that fail to adapt to new interaction contexts, as evidenced in Section \ref{sec:case}. To bridge imitation and on-policy agentic exploration, Stage 3 introduces a reasoning-focused RL phase that encourages adaptive reasoning before multi-turn RL.

We adopt GRPO~\cite{shao2024deepseekmath} for both reinforcement stages of \birdrl. In Stage 3, following the single-turn output format $y=(t,\sigma)$, the policy samples a group of $N$ candidate outputs $\{y_i=(t_i,\sigma_i)\}_{i=1}^{N}$ for each input $x$. Each solution $\sigma_i$ is evaluated by the execution-based reward defined below, and the resulting group-relative advantages are used to optimize:
\begin{equation}
\label{eq:6}
\mathcal{J}_{\mathrm{GRPO}}(\theta)
= \mathbb{E}\left[ \frac{1}{N} \sum_{i=1}^{N} 
\min \left(
\rho_i A_i,\;
\mathrm{clip}(\rho_i, 1-\epsilon, 1+\epsilon)A_i
\right)
\right]
- \beta D_{\mathrm{KL}}(\pi_\theta \,\|\, \pi_{\mathrm{ref}}),
\end{equation}
where $\rho_i=\frac{\pi_\theta(y_i \mid x)}{\pi_{\theta_{\mathrm{old}}}(y_i \mid x)}$ is the likelihood ratio, $A_i$ is the advantage, and $\pi_{\mathrm{ref}}$ is initialized from the Stage 2 policy.

\paragraph{Reward Design.}
Since reasoning quality is difficult to verify directly, we supervise it through the functional correctness of the generated SQL. Given the test suite $\mathcal{C}$, the reward is defined as:
\begin{equation}
R = \frac{1}{|\mathcal{C}|} \sum_{c \in \mathcal{C}} 
\mathbb{I}\big(c(\sigma)=\texttt{True}\big),
\label{eq:7}
\end{equation}
where $\mathbb{I}(\cdot)$ is the indicator function. A solution receives credit for each test case it passes. The reward corresponds to the fraction of test cases passed by the generated SQL.

\subsection{Stage 4: Stateful Interactive Agentic Reinforcement}
\label{sec:stage4}

\paragraph{Agentic RL Training Objective.} As the culmination of the training curriculum, Stage 4 aims to enable autonomous multi-turn exploration with tool augmentation through agentic RL training. For each input task $x$, a group of $N$ multi-turn trajectories $\{\tau_i\}_{i=1}^{N}$ is sampled. The final solution $\sigma_i$ is determined by the \texttt{submit\_solution} action when available, or otherwise by the last generated query. The optimization follows the GRPO objective in Eq.~(\ref{eq:6}), formulated at the trajectory level. Specifically, the likelihood ratio is computed over the full trajectory $\tau_i$ instead of a single-turn output $y_i$, and the reference policy $\pi_{\mathrm{ref}}$ is updated to the policy obtained from Stage 3.

\paragraph{Reward Design.} 
The objective of the final stage is to produce a functionally correct solution through iterative tool-augmented interaction. While prior works \cite{pourreza2025reasoningsql, hua2026sqltrailmultiturnreinforcementlearning} typically augment verifiable execution rewards with elaborately designed proxy rewards, we observe that these proxy rewards can induce reward hacking in SQL debugging, as demonstrated in Section \ref{app:ablation}. To this end, we adhere to the execution-based reward defined in Eq.(\ref{eq:7}) assigned exclusively to the final solution, providing an unbiased signal directly aligned with the stage goal. To encourage unconstrained exploration, we apply no process-level rewards. Eliminating the step-wise credit assignment prevents bias toward specific interactions and purely aligns optimization with the final functional objective. Detailed designs of the proxy-reward variants used for comparison are provided in Appendix \ref{app:reward}.

\section{Trajectory-Scoped Persistent Execution Infrastructure}
\label{sec:infra}

As discussed in Section~\ref{sec:tool-centric}, standard tool-centric loops record tool observations in the textual history but do not explicitly bind a rollout to a persistent execution environment. This design is insufficient for stateful SQL debugging, where each trajectory must retain the database effects of its earlier actions for subsequent hypothesis testing. To support scalable stateful RL with environment isolation, \birdrl introduces a Trajectory-scoped Persistent Execution Infrastructure, as shown in Figure~\ref{fig:main_fig} (right). It consists of two main modules: a worker-partitioned database pool (Section \ref{sec:database_pool}) for reusable storage and a customized stateful agent loop \ref{sec:agent_loop} for trajectory-level environment binding.

\subsection{Worker-Partitioned Database Pool}
\label{sec:database_pool}

The database pool maintains a pre-allocated set of reusable database replicas to avoid repeatedly instantiating a fresh database for every rollout. Given template databases $\mathcal{D}=\{d_1,\ldots,d_N\}$, the pool maintains $K$ working replicas for each database $d_i$, denoted as $\{d_i^{(1)},\ldots,d_i^{(K)}\}$. Since replica availability is managed locally by distributed rollout workers, replicas are partitioned across $W$ workers:
\begin{equation}
\{ d_i^{(1)}, \dots, d_i^{(K)} \} =
\bigcup_{w=1}^{W} \mathcal{E}_i^{(w)}, \quad
\mathcal{E}_i^{(w)} \cap \mathcal{E}_i^{(w')} = \emptyset \; (w \ne w'),
\end{equation}
where $\mathcal{E}_i^{(w)}$ denotes the worker-exclusive replica subset assigned to worker $w$. Each replica in the pool maintains two states, which are  \texttt{is\_available} for assignment availability and \texttt{need\_reset} for whether restoration is needed before reuse. This design enables efficient replica reuse while preventing cross-trajectory contamination.

\subsection{Customized Stateful Agent Loop}
\label{sec:agent_loop}
Standard tool-centric loops \cite{verl} usually dispatch each tool invocation as a request-scoped call: the execution backend receives the current tool input and returns an observation, while the request identifier only represents this individual call rather than the entire rollout. Such request-scope abstraction fails to preserve state continuity within a trajectory, which is essential for SQL debugging with state-mutating operations. We therefore make the environment management trajectory-scoped with four designs: 
\textbf{1) Stable Trajectory Identifier Binding:}
at the start of each rollout, we assign a stable trajectory identifier, and all subsequent \texttt{execute\_sql} calls within the same rollout use this identifier to retrieve the same database replica, ensuring intra-trajectory state persistence.
\textbf{2) Lightweight Instance Registry:}
we maintain a lightweight registry that maps each trajectory identifier to its allocated database replica, e.g., $(\texttt{trajectory\_id}, \texttt{worker\_id}, \texttt{db\_id})$. This registry serves as the assignment memory between the agent loop and the database pool, routing each tool call to the intended persistent environment.
\textbf{3) Deferred State Release:}
the assigned replica remains bound to the trajectory until termination, rather than being released after each tool call. It is normally released upon the terminal \texttt{submit\_solution} action, while an asynchronous timeout mechanism reclaims replicas from abnormal rollouts to prevent resource leakage.
\textbf{4) Dirty-Aware Lazy Reset:}
when a released replica has been modified, it is marked with \texttt{need\_reset=True} and returned to the pool without resetting at trajectory termination. The reset is deferred until the replica is allocated to a new trajectory, which avoids termination-phase lock contention while ensuring that every new rollout starts from a clean database state. To further clarify the workflow of the customized agent loop shown in Figure~\ref{fig:main_fig}, we present its pseudo-code in Algorithm~\ref{alg:workflow}.

\section{Experiments}

\subsection{Experiment Settings}

\paragraph{Datasets and Metrics.}
We evaluate \birdrl on heterogeneous SQL tasks covering both SQL debugging and SQL generation. For SQL debugging, we evaluate on \textsc{Bird-Critic-SQLite}~\cite{li2026swesql}, which contains 500 SQLite issues across query-like, management, and personalization categories, and \textsc{Bird-Critic-Multi}~\cite{li2026swesql}, which contains 570 issues across PostgreSQL, MySQL, SQL Server, and Oracle. For SQL generation, we evaluate on the \textsc{Bird-SQL} \cite{li2024bird} development set with 1,534 examples. We report Success Rate (SR) for SQL debugging and Execution Accuracy (EX) for SQL generation following the official metrics of each benchmark.

\paragraph{Implementation and Baselines.}
We implement \birdrl on Qwen2.5-Coder-7B and Qwen2.5-Coder-14B, producing the debugging-specialized \birdtalon models. We also train \birdzeno-7B as a unified model for both SQL debugging and generation through mixed-task training. All post-training stages are implemented with VERL~\cite{verl} and run on 8$\times$H100 GPUs. Our trajectory-scoped infrastructure is integrated as a plug-and-play patch compatible with general VERL-based rollout pipelines. For baseline comparison, we consider strong general-purpose LLMs, multi-turn SQL agents, and supervised debugging agents on SQL debugging, as well as recent RL-based text-to-SQL methods on SQL generation. Full dataset statistics, training details, inference settings, and compared-method descriptions are provided in Appendix~\ref{app:six_gym} and ~\ref{app:exp_details}.

\begin{table*}[t]
    \centering
    \caption{Success Rate (SR\,\%) of different models on
             \textsc{BIRD-CRITIC-SQLite}, evaluated under greedy decoding and grouped by issue category.
             \textbf{Bold numbers indicate the highest score in each column}, and
             \underline{underlined} numbers indicate the second highest.
             "Quer." refers to query-like issues, "Mana." refers to data-management issues,
             and "Pers." refers to personalized-function issues.}
    \label{tab:main_result}
    \setlength{\tabcolsep}{4pt}
    
    \renewcommand{\arraystretch}{0.95}
    \resizebox{0.8\textwidth}{!}{%
    \begin{tabular}{C{3.8cm}lcccc}
      \toprule
      \textbf{Type} & \textbf{Model} & \textbf{Quer.} & \textbf{Mana.} & \textbf{Pers.} & \textbf{Overall} \\
      \midrule
      \multirow{8}{*}{General-Purpose Models}
        & GPT-5.4-Pro       & 44.01 & 38.67 & 39.72 & 42.00 \\
        & Claude-Opus-4.6   & 50.35 & 44.00 & 39.01 & 46.20 \\
        & Claude-Sonnet-4.5 & 46.83 & 36.00 & 34.75 & 41.80 \\
        & Gemini-3.1-Pro    & \textbf{53.52} & 44.00 & \textbf{41.84} & \textbf{48.80} \\
        & GLM-4.7           & 48.24 & 40.00 & 33.33 & 42.80 \\
        & Kimi-K2.5         & 46.48 & 37.33 & 35.46 & 42.00 \\
        & MiniMax-M2-1      & 38.03 & 30.67 & 32.62 & 35.40 \\
        & Qwen3-Coder-480B  & 45.77 & 33.33 & 37.59 & 41.60 \\
      \midrule
      \multirow{5}{*}{Multi-Turn Agents}
        & Claude-Opus-4.6   & \textbf{53.52} & 42.67 & \underline{40.43} & \underline{48.20} \\
        & Claude-Sonnet-4.5 & 44.01 & \textbf{54.67} & 37.59 & 43.80 \\
        & Qwen3-Coder-480B  & 46.48 & 34.67 & 39.01 & 42.60 \\
        & BIRD-FIXER-7B     & 37.32 & 26.67 & 29.79 & 33.60 \\
        & BIRD-FIXER-14B    & 42.96 & 37.33 & 35.46 & 40.00 \\
      \midrule
      \multirow{2}{*}{Ours}
        & BIRD-Talon-7B     & 46.13 & 45.33 & \underline{40.43} & 44.40 \\
        & BIRD-Talon-14B    & \underline{51.06} & \underline{50.67} & \underline{40.43} & 48.00 \\
        
      \bottomrule
    \end{tabular}%
    }
\end{table*}

\begin{table}[t]
    \centering
    \caption{Detailed comparison of \birdtalon with other strong baselines on \textsc{BIRD-CRITIC-SQLite}. The SQL-ACT agent employs SQLs as actions. $\Delta$ denotes the absolute improvement of \birdtalon compared with the base model.}
    \label{tab:talon_scale}
    \small
    \setlength{\tabcolsep}{5pt}
    \resizebox{0.8\linewidth}{!}{%
    \begin{tabular}{lccccc}
        \toprule
        \textbf{Model} & \textbf{Base} & \textbf{\textsc{SQL-Act}} & \textbf{\textsc{Bird-Fixer}} & \textbf{\textsc{Bird-Talon}} & \textbf{$\Delta$ (\%)} \\
        \midrule
        Qwen-2.5-Coder-7B  & 26.20 & 29.80 &  33.60 & \textbf{44.40} & \textcolor{gainGreen}{+18.20}  \\
        Qwen-2.5-Coder-14B & 33.80 & 36.40 & 40.00 & \textbf{48.00} & \textcolor{gainGreen}{+14.20}\\
        \bottomrule
    \end{tabular}}
    \vspace{-0.8em}
\end{table}

\subsection{Overall Performance}
\paragraph{Results on SQL Debugging.}
Table \ref{tab:main_result} and Table \ref{tab:talon_scale} present the performance of \birdtalon comparing approaches with mainstream LLMs and advanced SQL debugging agents. We can observe that:

(1) \textbf{Considerable Challenge in Stateful SQL Issue Debugging:}
Even the strongest proprietary model reaches only 48.80\% SR, showing that SQL issue resolution over mutable databases is far from being solved. Category-level results further reveal that the difficulty mainly comes from state-dependent issues in which general-purpose models perform better on query-like cases, with 46.65\% average SR, but drop to 38.00\% and 36.79\% on management and personalization issues, respectively. In addition, multi-turn agentic inference improves average overall SR by only 3.87\%. This limited gain suggests that inference-time interaction alone is insufficient for robust stateful SQL debugging, motivating our agentic RL framework that enables adaptive reasoning through persistent database interaction. 

(2) \textbf{Superior Performance of \birdtalon:}
As shown in Table~\ref{tab:talon_scale}, \birdtalon brings consistent gains across both model scales, improving the Qwen2.5-Coder-7B and 14B backbones by \textbf{18.20\%} and \textbf{14.20\%}, respectively. We can also observe that directly applying the SQL-ACT scaffold produces only limited gains. While supervised trajectory imitation with \textsc{Bird-Fixer} improves performance further, it still falls substantially behind \birdtalon. This suggests that imitation alone is insufficient to capture the decision dynamics required for multi-turn SQL debugging. Instead, \birdrl enables the agent to adapt its reasoning and tool use through reasoning and agentic RL. Notably, \birdtalon-14B achieves \textbf{48.00\%} overall SR, approaching the comparable performance of the Claude-Opus-4.6 agent (48.20\%) while employing a much smaller open model, which is especially valuable for privacy-sensitive SQL applications.

\begin{table}[t]
\centering
\small
\setlength{\tabcolsep}{4pt}
\renewcommand{\arraystretch}{0.95}
\caption{Comparison of database lifecycle I/O during interactive rollout execution under trajectory-level isolation. “Emp. I/O” reports the measured online database lifecycle I/O from our implementation and excludes the one-time offline pool materialization cost.}
\label{tab:db-materialization}

\resizebox{0.7\linewidth}{!}{%
\begin{tabular}{c l l l}
\toprule
\textbf{Strategy}
& \textbf{Training-time I/O}
& \textbf{Emp. I/O ($\downarrow$)}
& \textbf{Cost Ratio} \\
\midrule
Per-rollout Container
& $O(NR\bar{S})$
& 2.94 TiB
& $12.8\times$ \\

Dynamic Reusable Pool
& \(O((B+q/k)R\bar{S})\)
& 236 GiB
& $1.0\times$ \\
\bottomrule
\end{tabular}
}
\end{table}

\subsection{Analysis of Database Reuse Efficiency.}

To validate our proposed pool-based container isolation strategy, we conduct an efficiency analysis of database lifecycle I/O under trajectory-level isolation. Unlike peak memory and storage capacity, which can be managed through resource provisioning and pool sizing, repeated database materialization and restoration place sustained pressure on storage bandwidth. The resulting I/O saturation can cause rollout stalls or timeouts. Such infrastructure-induced failures can distort the reward signal and destabilize on-policy optimization if recorded as zero-reward trajectories. Reducing this recurring overhead is therefore important for feasible and reliable agentic RL over resource-intensive database environments. 

As shown in Table~\ref{tab:db-materialization}, the training-time I/O column presents the theoretical online I/O complexity during rollout execution. The related parameter definition and detailed derivation of the complexity terms is provided in Appendix~\ref{app: complexity}. The naive per-rollout baseline scales with the total number of generated trajectories. In contrast, the dynamic reusable pool fundamentally amortizes the database instantiation cost via cross-rollout reusability. This strategic shift leaves the online I/O dominated mainly by resetting modified replicas to preserve clean trajectory-level isolation. Benefiting from the SFT stages in our framework, which effectively regularizes the agent's exploration behavior, only \textbf{9.25\%} of the practical trajectories execute state-altering queries and actually trigger a database reset. As a result, our proposed method guarantees a superior I/O efficiency compared to the naive baseline. 

However, dynamic online allocation is vulnerable to abrupt memory spikes under imbalanced database workloads, which can cause capacity overflows and destabilize rollout execution. We thus apply a pre-allocated database pool to mitigate this risk. Under the evaluated SQLite setting with batch size $B=16$ and rollout $R=16$, our design reduces the measured online lifecycle I/O by \boldmath{$12.8\times$} compared to per-rollout containers. The trade-off is an offline pool cost of 69 GiB, which is practically negligible relative to the terabyte-level massive online I/O savings. Notably, since this offline pool scales strictly with the number of unique databases rather than the massive training volume $N$, its storage overhead remains predictable and highly cost-effective even under large-scale trajectory-level isolation. As the first work to investigate pool-based environment isolation for data-centric tasks, our infrastructure enables stable and scalable stateful agentic RL training for SQL debugging, paving the way for more intelligent database environment management as future work.

\subsection{Results on Heterogeneous SQL Tasks.}

\paragraph{Overall Performance.} 
\begin{wraptable}{r}{0.50\columnwidth}
    \vspace{-1em}
    \centering
    \caption{Comparison of \birdzeno with strong RL-based generation-specialized methods. "Gen." = "Generation" and "Debug." = "Debugging".}
    \label{tab:sql_generation}
    \small
    \setlength{\tabcolsep}{4pt}
    \resizebox{0.7\linewidth}{!}{%
    \begin{tabular}{lcc}
        \toprule
        \textbf{Model} & \textbf{Gen.} &
        \textbf{Debug.} \\
        \midrule
        Think2SQL-7B       & 56.1 & 24.0 \\
        SQL-R1-7B       & 58.9 & 26.6 \\
        ReEX-SQL-7B        & 63.4 & - \\
        Reasoning-SQL-7B   & 64.0  & - \\
        Arctic-T2S-R1-7B   & \textbf{68.9}  & 29.8 \\
        SQL-Trail-7B       & 60.1  & - \\
        \midrule
        \textbf{BIRD-Zeno-7B} (Ours)      & 63.9 & \textbf{44.6} \\
        \bottomrule
    \end{tabular}}
    \vspace{-1.0em}
\end{wraptable}

Table \ref{tab:sql_generation} reports the performance of \birdzeno, which is trained under the data-mixed \birdrl scaffold, on heterogeneous SQL tasks across SQL generation and issue resolution.  \birdzeno-7B maintains superior debugging capabilities while achieving results comparable to strong RL-based generation-specific specialists on the \textsc{BIRD-SQL} development set. To our knowledge, this study is the first exploration of unified training in the SQL domain, highlighting the potential for multi-objective optimization within a single framework across real-world data-centric tasks.

\paragraph{Data Mixture Ratio Discussion.}
\label{sec:data_mixing}
The data mixture for multi-task training is a complex trade-off between task-specific performance and cross-task interference \cite{data_mix}. To estimate a suitable mixture ratio for \birdzeno, we conduct a controlled study with Qwen2.5-Coder-7B under three debugging-generation ratios. Specifically, we maintain a fixed total training set of 4,000 instances and evaluate the following settings: \textit{debug-heavy (3:1)}, \textit{balanced (1:1)}, and \textit{generation-heavy (1:3)}. For efficiency, SQL generation is evaluated on \textsc{Bird-Mini} \cite{birdmini}, a mini subset of the BIRD development set.

\begin{wrapfigure}{r}{0.45\textwidth}
\vspace{-0.4cm}
\centering
\includegraphics[width=0.7\linewidth]{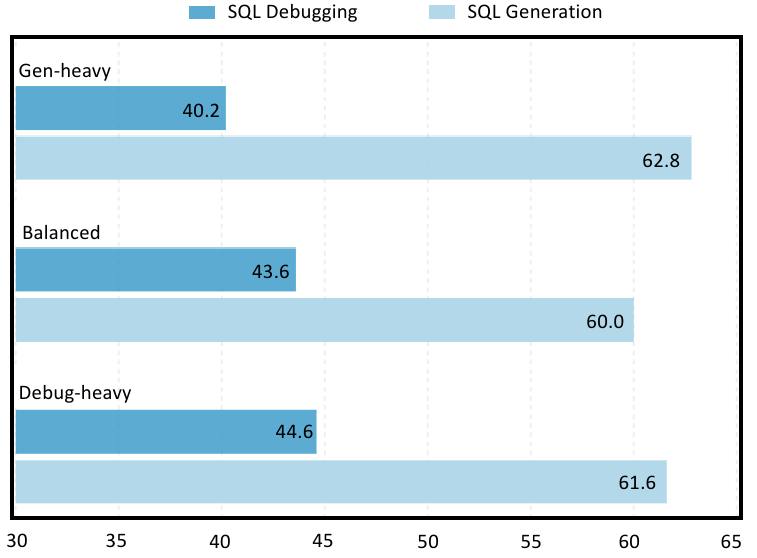}
\caption{Effect of data mixture ratio.}
\label{fig:data_mix}
\vspace{-0.3cm}
\end{wrapfigure}

  As illustrated in Figure \ref{fig:data_mix}, varying the mixture ratio reveals a clear trade-off between SQL generation and issue resolution. We speculate that this asymmetry arises because the two data sources play different roles within the shared execution-grounded reward space. Generation data primarily establishes schema grounding and compositional SQL priors, whereas debugging data shapes exploration behavior for effective information acquisition from databases. In light of this trade-off, we adopt the debug-heavy ratio of 3:1 as the final mixture for \birdzeno, which yields the strongest issue-resolution performance while retaining competitive SQL generation capability. 

\subsection{Case Study: Effectiveness of the Adaptive Reasoning Reinforcement Stage}

\label{sec:case}

\begin{table}[t]
\centering
\scriptsize
\setlength{\tabcolsep}{3pt}
\renewcommand{\arraystretch}{1.18}
\caption{Case study illustrating the effectiveness of Stage 3, with the full trajectory shown in Figure~\ref{fig:case}.}
\label{tab:case}

\begin{minipage}{0.9\linewidth}
\begin{tabularx}{\linewidth}{p{0.16\linewidth} X X}
\toprule
\textbf{Setting} & \textbf{Reasoning Trace} & \textbf{Action Trace} \\
\midrule
w/o Stage 3 &
\emph{``... the problematic query joins budget and expense ... To get the correct total budget per event, I need to aggregate budget amounts at the event level using MAX or MIN, \bad{since each event has one budget} ...''} &
\texttt{SELECT ... SUM(expense.cost), \bad{MAX(budget.amount) AS total\_budget} ... GROUP BY budget.link\_to ...} \\
\midrule
w/ Stage 3 (Ours) &
\emph{``... \good{there are multiple budget lines} ..., and \good{each budget line has multiple expense records} ... The solution is to \good{pre-aggregate expenses per budget line first}, then join with budget ...''} &
\texttt{SELECT SUM(sub\_exp.total\_cost), SUM(budget.amount) ... FROM (\good{SELECT link\_to\_budget, SUM(cost) ... GROUP BY link\_to\_budget}) sub\_exp ...} \\
\bottomrule
\end{tabularx}
\end{minipage}
\end{table}

Table \ref{tab:case} presents a concrete case illustrating the effectiveness of our Adaptive Reasoning Reinforcement Stage (Stage 3), with the full trajectories provided in Appendix~\ref{app:case} (Figure~\ref{fig:case}). In this case, both agents correctly identify the root cause as aggregation errors in the earlier step, demonstrating the effectiveness of our first two training stages. However, their solutions later diverge: 1) \textbf{w/o Stage 3:} The agent fails to incorporate feedback and instead produces \textbf{\textit{independent}} step-wise erroneous reasoning, which is a typical side effect of SFT. 2) \textbf{w/ Stage 3:} RL enables the agent to leverage the context \textbf{\textit{coherently}} to apply the knowledge acquired during SFT, ultimately resulting in the correct reasoning and actions. This case underscores a critical shift from memorized debugging patterns toward adaptive, evidence-grounded reasoning, proving the necessity of Stage 3.

\subsection{Ablation Study}
\label{app:ablation}

Table \ref{tab:ablation} presents our ablation study, with fine-grained breakdowns across issue categories. The removal of the training stages shown in (a)-(c) leads to obvious performance drops, indicating that the overall gain arises from the progressive curriculum. This pattern underscores the core design principle of \birdrl: imitation first anchors the policy in structured diagnostic reasoning and coherent tool use, while RL subsequently transforms into a more autonomous policy capable of feedback-driven exploration beyond the demonstrated trajectories. Notably, the effect of excluding the agentic RL stage (Table \ref{tab:ablation} (a)) is more pronounced on management ($\downarrow$9.33) and personalization issues ($\downarrow$7.10) than on query-like cases ($\downarrow$0.33), suggesting that agentic RL is particularly beneficial for resolutions that require stateful hypothesis testing and iterative revision.

Table \ref{tab:ablation} (d)-(f) compares the purely execution-based reward employed in the agentic RL stage with three variants incorporating auxiliary proxy reward signals. Specifically, variant (d) applies rule-based partial-correctness signals to the submitted solution, whereas variants (e) and (f) additionally introduce trajectory-level supervision, scoring both the intermediate steps and the solution via an LLM judge and heuristic rubric criteria, respectively. Detailed reward designs are provided in Appendix \ref{app:reward}. The superior performance of the sparse, purely execution-based reward indicates that the reward fidelity outweighs density in SQL debugging tasks. Given the extreme error sensitivity of the task, proxy rewards are prone to reward hacking, steering the policy toward superficially plausible syntax or favorable trajectories instead of genuine functional resolution. This suggests that heuristic proxy rewards, while widely used, often require substantial expert effort to construct. To this end, \birdrl adopts the neat but effective execution-based design while remaining extensible to future sophisticated reward functions to further improve the upper bound of RL in data-centric tasks.

\begin{table}[t]
    \centering
    \caption{Ablation study on the training curriculum and reward design of \birdrl.}
    \label{tab:ablation}
    \small
    \setlength{\tabcolsep}{4pt}
    \renewcommand{\arraystretch}{1.03}
    \begin{tabular}{@{}cl@{\hspace{0.75em}}cccc@{}}
        \toprule
        \multirow{2}{*}{\textbf{\textsc{Type}}}
        & \textbf{\textsc{Method}} & \textbf{\textsc{Quer.}} & \textbf{\textsc{Mana.}} & \textbf{\textsc{Pers.}} & \textbf{\textsc{Overall}} \\
        \cmidrule(l){2-6}
        & \textbf{BIRD-Talon-7B} & 46.13 & 45.33 & 40.43 & \textbf{44.40} \\
        \midrule
        \multirow{3}{*}{Training Strategy}
        & (a) \textit{Up to Reasoning RL (Stages 1-3)} & 45.77 & 36.00 & 33.33 & \textbf{40.60}\,\raisebox{0.2ex}{\tiny($\downarrow$3.80)} \\
        & (b) \textit{Up to Tool-Use Imitation (Stages 1-2)} & 45.07 & 33.33 & 30.50 & \textbf{39.20}\,\raisebox{0.2ex}{\tiny($\downarrow$5.20)} \\
        & (c) \textit{Reasoning Imitation Only (Stage 1)} & 41.55 & 28.00 & 26.24 & \textbf{35.20}\,\raisebox{0.2ex}{\tiny($\downarrow$9.20)} \\
        \midrule
        \multirow{3}{*}{Reward Design}
        & (d) \textit{w/ Partial Correctness} & 45.77 & 41.33 & 37.59 & \textbf{42.80}\,\raisebox{0.2ex}{\tiny($\downarrow$1.60)} \\
        & (e) \textit{w/ LLM-as-a-Judge} & 45.77 & 38.67 & 34.04 & \textbf{41.40}\,\raisebox{0.2ex}{\tiny($\downarrow$3.00)} \\
        & (f) \textit{w/ Heuristic Rubric} & 45.42 & 40.00 & 36.17 & \textbf{42.00}\,\raisebox{0.2ex}{\tiny($\downarrow$2.40)} \\
        \bottomrule
    \end{tabular}
\end{table}

\section{Related Work}
\paragraph{LLMs for Text-to-SQL.} In recent years, the application of LLMs has improved the performance of text-to-SQL to another level of intelligence~\cite{geoquery, yaghmazadeh2017sqlizer, t2s_survey, huo2025micro}. Early research primarily leveraged the in-context learning capabilities of LLMs to build text-to-SQL systems through sophisticated prompt engineering~\cite{gao2023text, pourreza2023din, gao2023text, qu2024tasql, pourreza2025chasesql}. Driven by the advent of language agents, recent text-to-SQL research has pivoted toward multi-agent architectures~\cite{talaei2024chess, macsql-2025, chaturvedi2025sqlofthought}. A critical component emerging from these architectures is the automatic error correction mechanism. Although recent agents increasingly incorporate automatic correction loops to improve response accuracy~\cite{shinn2023reflexion, madaan2023selfrefine, pourreza2025chasesql}, the distinct challenge of debugging the full spectrum of SQL operations for real-world database administration remains largely unexplored. Our work is expressly designed to bridge this gap.

\paragraph{RL for Reasoning.} Reinforcement learning (RL) has recently emerged as a promising paradigm for improving the reasoning capabilities of LLMs, enabling models to autonomously learn effective policies with the guidance of reward signals~\cite{guo2025deepseek, ouyang2022training}. It has shown impressive success on reasoning-intensive tasks such as mathematical solving~\cite{shao2024deepseekmath}, code generation~\cite{liu2023rltf, shojaee2023executionbased}, and visual reasoning~\cite{team2026kimi, sun2024aligning, yu2025rlaif}. In light of these advancements, recent research in the text-to-SQL domain has started to leverage RL-based frameworks to steer the query synthesis process~\cite{papicchio2025think2sql, yao2025arctic, ma2026sqlr, dai2025reex, pourreza2025reasoningsql, hua2026sqltrailmultiturnreinforcementlearning}. Our work aims to extend this line of research beyond static one-shot SQL synthesis to dynamic, stateful SQL issue resolution, where the agent must learn to diagnose and revise candidate repairs through multi-step interaction.

\section{Conclusion}
In this work, we introduce \birdrl, a four-stage agentic RL curriculum for SQL issue resolution. To enable scalable training in data-centric environments, we equip \birdrl with a Trajectory-scoped Persistent Execution Infrastructure that supports isolated stateful interactions through orchestrated database-replica reuse. Built on this framework, we release the \birdtalon model family for SQL debugging. We further introduce \birdzeno, a unified multi-task model trained with a principled data-mixing strategy, to examine the generality of our curriculum beyond single-objective optimization. Together, these contributions establish \birdrl as a scalable framework for data-centric agent training, paving the way for agentic RL in heavy, stateful database environments and highlighting the potential of multi-objective training for real-world SQL applications.

\newpage
\section{Acknowledgments}

We thank the anonymous reviewers and committees for their helpful comments, suggestions and organizations. Reynold Cheng, Ge Qu, Jinyang Li, Xiaohan Xu, Xiaolong Li, Nan Huo, and Ziwei Tang are supported by the Research Grant Council of Hong Kong (RGC Project HKU 17202325) and the University of Hong Kong (Project 2609260682). Ge Qu and Jinyang Li are supported by HKU Presidential PhD Scholar Programme. Ge Qu is also funded by Hong Kong PhD Fellowship Scheme.

\bibliographystyle{plainnat}
\bibliography{neurips_2026}

\newpage
\appendix

\section{\textsc{SIX-GYM-SQLite}: Training Environment Construction}
\label{app:six_gym}
Training LLMs as autonomous agents for complex tasks fundamentally requires an integrated environment that combines high-quality task suites, executable environments, and rigorous test verification. To efficiently support the massive trial-and-error exploration inherent in optimizing these agents via RL, with server-based database systems such as PostgreSQL and MySQL, the lightweight, in-process nature of SQLite eliminates the client-server overhead of engines like PostgreSQL, enabling the high-throughput parallel rollouts essential for large-scale SQL agent training. While the existing SQL debugging benchmark SWE-SQL has introduced SIX-GYM environment for PostgreSQL, it lacks a dedicated training environment specifically constructed for SQLite. To address this gap, we contribute SIX-GYM-SQLite as a practical resource extension of the prior work, employing it as the underlying sandbox to automatically synthesize the stage-specific training data for \birdrl.

\subsection{SIX-GYM-SQLite}
Following the SQL-Rewind methodology introduced in SWE-SQL, we construct SIX-GYM-SQLite by adapting the original SIX-GYM training environment to SQLite. The resulting task suite combines instances adapted from the original SIX-GYM task suites with additional instances synthesized from newly collected Stack Overflow posts. We retain 5,000 high-quality instances based on answer quality, SQL complexity, and taxonomic coverage.

To prevent training–evaluation contamination, the SQLite adaptation preserves the database- and source-level separation established by the SWE-SQL construction protocol. Specifically, SIX-GYM-SQLite and BIRD-CRITIC-SQLite retain the database schemas of their respective counterparts in SWE-SQL, which uses disjoint database domains and instances for training and evaluation. At the source level, the inherited construction protocol excludes Stack Overflow posts used in BIRD-CRITIC from the training source pool by post ID prior to synthesis. During subsequent supervision generation for Stages 1 and 2, the teacher model is provided only with training instances and their reference solutions, without access to evaluation instances.

Furthermore, we exhaustively compared all 500$\times$5,000 train–test pairs using a dual-view TF-IDF audit over normalized issue descriptions and buggy SQL. A pair was flagged as a high-confidence near-duplicate only when the same training instance exceeded 0.85 text similarity and 0.90 SQL similarity. Only 3 of 500 tasks (0.6\%) are highly similar to training tasks, and all three are permission-privilege-related tasks, a category that is inherently shared, since privilege management (e.g., \texttt{GRANT/REVOKE} and access-control patterns) is core database knowledge that neither training nor evaluation can meaningfully exclude. Excluding these three flagged evaluation tasks leaves our main conclusions unchanged.

\begin{figure}[t]
    \centering
    \begin{subfigure}[t]{0.48\linewidth}
        \centering
        \includegraphics[width=\linewidth]{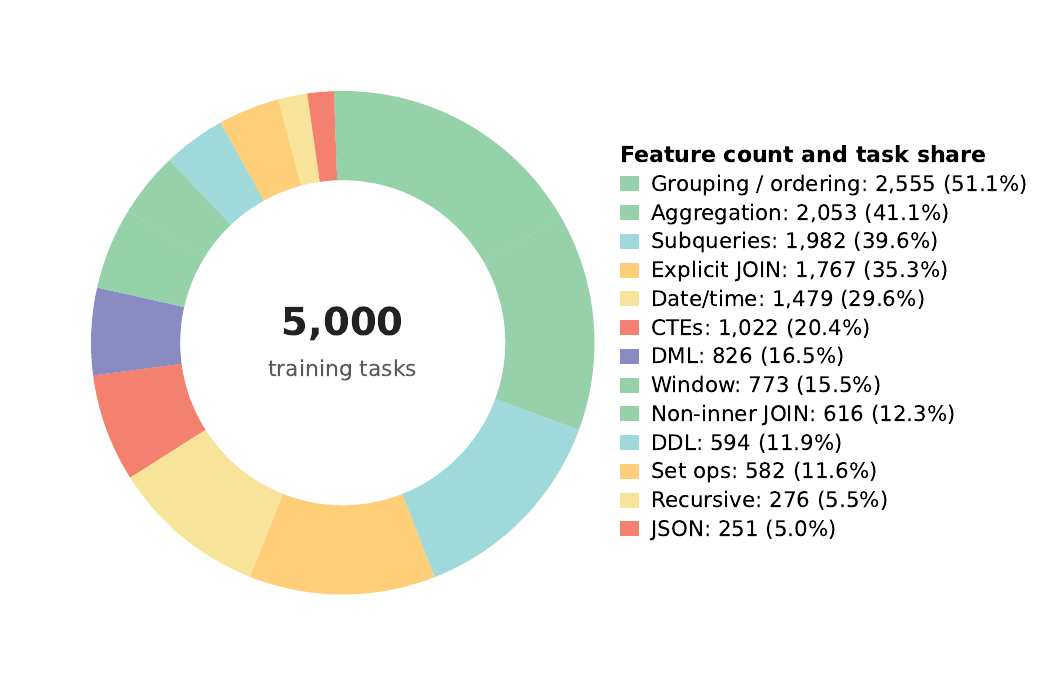}
        \caption{SQL usage categories. Categories are non-exclusive; percentages indicate the share of tasks containing each feature.}
        \label{fig:six_gym_sql_feature_distribution}
    \end{subfigure}
    \hfill
    \begin{subfigure}[t]{0.48\linewidth}
        \centering
        \includegraphics[width=\linewidth]{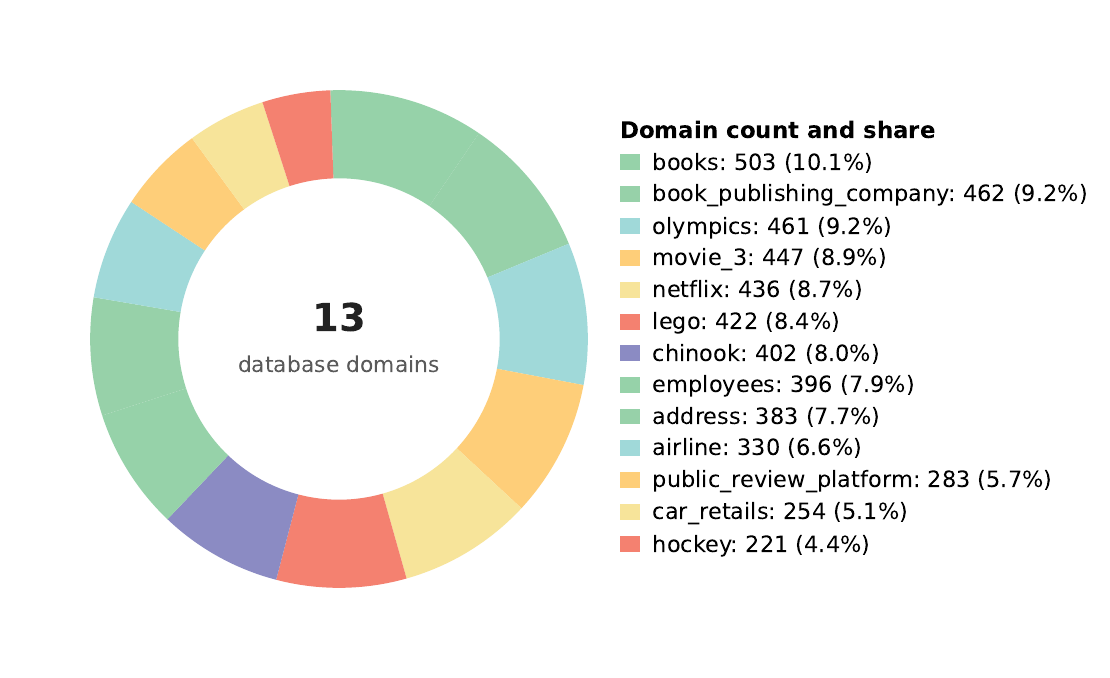}
        \caption{Database domains. SIX-GYM-SQLite spans 5,000 tasks across 13 SQLite database domains.}
        \label{fig:six_gym_training_domains}
    \end{subfigure}
    \caption{Statistics of SIX-GYM-SQLite training data, covering SQL feature usage and database-domain distribution.}
    \label{fig:six_gym_statistics}
\end{figure}

Table~\ref{tab:six_gym_data_characteristics} summarizes the overall data characteristics, including task categories, token lengths, evaluation scripts, and database coverage.
Table~\ref{tab:six_gym_category_statistics} further breaks down query, issue SQL, solution SQL, and test-case lengths by task category.
Figure~\ref{fig:six_gym_statistics} visualizes the distribution of SQL usage categories and database domains, showing that SIX-GYM-SQLite covers diverse SQL operations and application contexts.

\begin{table}[t]
\centering
\caption{Data characteristics of SIX-GYM-SQLite. Length statistics are reported as mean/max token counts using the \texttt{cl100k\_base} tokenizer.}
\label{tab:six_gym_data_characteristics}
\begin{tabular}{lc}
\toprule
Statistic & SIX-GYM-SQLite \\
\midrule
Total Issues & 5,000 \\
\# of query-like issues & 3,094 \\
\# of management issues & 1,087 \\
\# of personalization issues & 819 \\
user query length (mean/max) & 162.55/917 \\
issue SQL length (mean/max) & 85.50/898 \\
solution SQL length (mean/max) & 97.66/1,976 \\
test-case length (mean/max) & 221.38/1,601 \\
\# distinct test cases & 2,477 \\
\# preprocess SQLs & 3,512 \\
\# clean\_up SQLs & 969 \\
\# multi-SQL issue instances & 454 \\
\# database domains & 13 \\
\bottomrule
\end{tabular}
\end{table}

\begin{table}[t]
\centering
\caption{SIX-GYM-SQLite statistics grouped by task category. Length statistics are reported as mean/max token counts.}
\label{tab:six_gym_category_statistics}
\resizebox{0.9\linewidth}{!}{%
\begin{tabular}{lccccc}
\toprule
Category & Count & User Query & Issue SQL & Solution SQL & Test Cases \\
\midrule
Query & 3,094 & 183.93/917 & 86.76/751 & 96.22/1,976 & 143.74/1,601 \\
Management & 1,087 & 120.20/535 & 73.51/802 & 83.84/802 & 446.16/1,085 \\
Personalization & 819 & 138.01/421 & 96.63/898 & 121.40/1,041 & 216.33/1,418 \\
Overall & 5,000 & 162.55/917 & 85.50/898 & 97.66/1,976 & 221.38/1,601 \\
\bottomrule
\end{tabular}%
}
\end{table}

\section{Methodology Details}
\subsection{Workflow Orchestration of the Customized Stateful Agent Loop}

\label{app:pseudo}

Algorithm~\ref{alg:workflow} presents the pseudo-code of the customized stateful agent loop, illustrating how trajectory-level database binding, dirty-aware lazy reset, and replica release are orchestrated during rollout execution.

\begin{algorithm}[t]
\caption{Workflow Orchestration Pseudo Code}
\label{alg:workflow}
\begin{algorithmic}[1]
\Require Task instance $x$, target database template $d$, worker-partitioned DB pool $\mathcal{P}$
\Ensure Final submitted solution or timeout termination

\State Generate a unique trajectory identifier $\mathrm{traj\_id}$
\State Request an available replica $r$ of database $d$ from the local worker pool $\mathcal{P}$
\If{no replica is immediately available}
    \State Wait in the allocation queue until a replica is released or timeout occurs
\EndIf
\If{allocation timeout occurs}
    \State \Return timeout failure
\EndIf
\State Set $r.\texttt{is\_available} \leftarrow \texttt{False}$
\State Register $(\mathrm{traj\_id}, \mathrm{worker\_id}, \mathrm{db\_id})$ in the stateful agent loop
\If{$r.\texttt{need\_reset} = \texttt{True}$}
    \State Restore $r$ from the clean database template
    \State Set $r.\texttt{need\_reset} \leftarrow \texttt{False}$
\EndIf

\While{the trajectory has not terminated}
    \State The agent generates an action $a$
    \If{$a$ is \texttt{execute\_sql}}
        \State Lookup $(\mathrm{worker\_id}, \mathrm{db\_id})$ by $\mathrm{traj\_id}$ in the stateful agent loop
        \State Execute the SQL action on the trajectory-bound replica $r$
        \State Return the database observation to the agent
        \If{the SQL action mutates the database state}
            \State Set $r.\texttt{need\_reset} \leftarrow \texttt{True}$
        \EndIf
    \ElsIf{$a$ is \texttt{submit\_solution}}
        \State Record the submitted solution
        \State Break
    \EndIf
\EndWhile

\State Release $r$ back to the local worker pool $\mathcal{P}$
\State Set $r.\texttt{is\_available} \leftarrow \texttt{True}$
\State \Return submitted solution or timeout result
\end{algorithmic}
\end{algorithm}

\subsection{Auxiliary Proxy Reward Designs}
\label{app:reward}
In this section, we detail the auxiliary proxy-reward variants used in the ablation study. These variants augment the final execution-based reward with additional signals intended to provide denser supervision during agentic RL. Specifically, we consider three representative designs: partial-correctness rewards, LLM-as-a-judge trajectory scoring, and heuristic rubric-based rewards. All these rewards are used as auxiliary shaping signals on top of the execution-based reward. They are activated only for trajectories whose execution test pass rate is zero, so that any trajectory passing at least one test case is still primarily optimized by the execution-based reward.

\paragraph{Partial Correctness Reward.}
For zero-pass trajectories, we provide two lightweight partial signals: one for syntactic correctness and one for schema linking. The first signal rewards SQL that is syntactically valid and executable, while the second measures whether the generated SQL refers to schema elements consistent with the gold SQL.

Formally, the additional partial reward is defined as:
\begin{equation}
R_{\mathrm{partial}}
=
0.1 \times \mathbb{I}_{\mathrm{exec}}
+
0.1 \times F_{1}(s_{p}, s_{g})
\end{equation}
where $\mathbb{I}_{\mathrm{exec}}$ indicates whether the submitted SQL is syntactically valid and executable. $s_{p}$ and $s_{g}$ denote the sets of schema items appearing in the predicted and gold SQL queries, respectively. The schema-linking score is computed as:
\begin{equation}
F_{1}(s_{p}, s_{g})
=
\frac{2 \times P(s_{p}, s_{g}) \times R(s_{p}, s_{g})}
{P(s_{p}, s_{g}) + R(s_{p}, s_{g})},
\end{equation}
with
\begin{equation}
P(s_{p}, s_{g})
=
\frac{|s_{p} \cap s_{g}|}{|s_{p}|},
\quad
R(s_{p}, s_{g})
=
\frac{|s_{p} \cap s_{g}|}{|s_{g}|}.
\end{equation}

\paragraph{LLM-as-a-judge trajectory scoring}
The LLM-as-a-Judge reward uses strong proprietary LLMs as the judge model to score both the debugging trajectory and the final SQL solution using the prompt shown in Figure \ref{fig:llm_as_judge_prompt_1} and \ref{fig:llm_as_judge_prompt_2}. The judge assigns a trajectory quality score $q_{\mathrm{traj}}\in[0,5]$ and a solution quality score $q_{\mathrm{sol}}\in[0,10]$, with the maximum total score denoted as $Q_{\max}=15$. Since these trajectories have already failed execution tests, the auxiliary final score is computed by:

\begin{equation}
R_{\mathrm{judge}}
=
0.2 \times
\frac{q_{\mathrm{traj}} + q_{\mathrm{sol}}}{Q_{max}}.
\end{equation}

\paragraph{Heuristic Rubric-Based Reward.}
The heuristic rubric-based reward consists of two criteria: 1) whether the submitted SQL executes successfully, and 2) whether the model identifies known bugs in its reasoning. The bug-identification criterion is computed by matching the model's reasoning against a pre-extracted bug checklist with an LLM verifier. The reward is then computed as

\begin{equation}
R_{\mathrm{rubric}}
=
0.1 \times \mathbb{I}_{\mathrm{exec}}
+
0.1 \times
\frac{N_{\mathrm{bug}}^{\mathrm{id}}}{N_{\mathrm{bug}}}.
\end{equation}

Here, $\mathbb{I}_{\mathrm{exec}}$ indicates whether the submitted SQL executes without syntax errors, $N_{\mathrm{bug}}$ is the number of known bugs in the checklist, and $N_{\mathrm{bug}}^{\mathrm{id}}$ is the number of bugs identified by the model. The prompts used for bug extraction and verification in this process are presented in Figure \ref{fig:rubric_extract_prompt} and \ref{fig:rubric_verify_prompt}.

\section{Experiment Details}
\label{app:exp_details}

This section summarizes the implementation details, and compared methods used in our experiments.

\subsection{Alias of LLMs}
\label{app:alias}
The following aliases are used for the models in this work:
\begin{itemize}
    \item \textbf{GPT-5.4-Pro}: \texttt{gpt-5.4-pro}
    \item \textbf{Claude-Opus-4.6}: \texttt{claude-opus-4.6}
    \item \textbf{Claude-Sonnet-4.5}: \texttt{claude-sonnet-4-5-20250929}
    \item \textbf{Gemini-3.1-Pro}: \texttt{gemini-3.1-pro-preview}
    \item \textbf{GLM-4.7}: \texttt{glm-4.7}
    \item \textbf{Kimi-K2.5}: \texttt{kimi-k2.5}
    \item \textbf{MiniMax-M2-1}: \texttt{minimax-m2-1}
    \item \textbf{Qwen3-Coder-480B}: \texttt{Qwen3-Coder-480B}
\end{itemize}

All open-source models are downloaded from Hugging Face\footnote{\url{https://huggingface.co/}}:
\begin{itemize}
    \item \textbf{Qwen-Coder}: \texttt{Qwen2.5-Coder-7B-Instruct}, \texttt{Qwen2.5-Coder-14B-Instruct}
    \item \textbf{SQL generation baselines}: \texttt{Think2SQL-7B}, \texttt{SQL-R1-7B}, \texttt{Arctic-Text2SQL-R1-7B}
\end{itemize}

\subsection{Model Implementation Details}
\label{app:model_imp}

We post-train all our models using the VERL framework \cite{verl} (v0.7.0.dev0)\footnote{\url{https://verl.readthedocs.io/en/latest/index.html}} and vLLM \cite{kwon2023efficient} (v0.11.0)\footnote{\url{https://docs.vllm.ai/en/latest}}. All experiments are conducted on 8$\times$H100 GPUs with 80GB memory.
 Our training pipeline has four stages. The first two stages are supervised fine-tuning: Stage 1 trains single-turn SQL reasoning for 3 epochs with batch size 16, maximum length 8192, and learning rate 1e-5; Stage 2 trains multi-turn tool-use trajectories for 3 epochs with batch size 16, maximum length 16000, and learning rate 1e-5. The last two stages use GRPO \cite{shao2024deepseekmath}. Stage 3 performs single-turn reasoning RL with rollout size 16, batch size 16, PPO mini-batch size 16, learning rate 1e-6, KL coefficient 0.01, maximum prompt length 15000, and maximum response length 5000. Stage 4 performs multi-turn agentic RL with asynchronous vLLM rollouts up to 5 assistant turns. Unless otherwise specified, it follows the same RL hyperparameters as Stage 3, except that the rollout size is set to 4. The specific training hours for each backbone model are shown in Table \ref{tab:hours}. During inference, we use vLLM with greedy decoding (temperature 0) and a maximum input length of 15000 tokens. We report experimental results as the average of five repeated trials.

\begin{table}[t]
  \caption{GPU hours spent to train each backbone model.}
  \small
  \centering
  \resizebox{0.4\hsize}{!}{\begin{tabular}{cc}
    \toprule
    \textbf{Model} & \textbf{GPU Hours} \\
    \midrule
    BIRD-Talon-7B & 93.38  \\
    BIRD-Zeno-7B & 80.36 \\
    BIRD-Talon-14B & 152.75 \\
    \bottomrule
  \end{tabular}}
  \label{tab:hours}
\end{table}

For proprietary-model inference, we use the official API providers, including OpenAI (\url{https://openai.com/}), Anthropic (\url{https://www.anthropic.com/}), and Google (\url{https://gemini.google.com/}). All open-source models are downloaded from Hugging Face\footnote{\url{https://huggingface.co/}}; the Qwen-Coder backbones refer to \texttt{Qwen2.5-Coder-7B-Instruct} and \texttt{Qwen2.5-Coder-14B-Instruct}. 

\subsection{Compared Methods}
\label{app:compared_methods}
This section provides additional details of the compared methods reported in Tables~\ref{tab:main_result}, \ref{tab:talon_scale}, and \ref{tab:sql_generation}.

\paragraph{SQL debugging baselines.}
The General-Purpose Models group evaluates frontier proprietary models and strong open-weight models under a single-turn prompt-based SQL repair setting. Given the user issue description, database schema, and buggy SQL query, the model is directly prompted to produce the corrected SQL solution without interacting with the database during inference. This setting follows the baseline SQL debugging prompt shown in Figure~\ref{fig:sql_debug_baseline_prompt}. We include GPT-5.4-Pro, Claude-Opus-4.6, Claude-Sonnet-4.5, Gemini-3.1-Pro, GLM-4.7, Kimi-K2.5, MiniMax-M2-1, and Qwen3-Coder-480B to establish broad baselines on \textsc{Bird-Critic-SQLite}.

The Multi-Turn Agents group evaluates models with database interaction through the \textsc{SQL-Act} scaffold. Following the ReAct paradigm~\cite{yao2023react}, \textsc{SQL-Act} interleaves reasoning, SQL actions, and database observations over multiple turns. At each step, the model may execute an arbitrary SQL statement against the database environment and use the returned observation to refine its diagnosis. The trajectory continues until the model submits a final repaired SQL solution. Compared with tool-based agents whose action space is restricted to predefined operations such as schema inspection or sample retrieval, \textsc{SQL-Act} treats executable SQL itself as the action space, enabling more flexible debugging strategies. The system and user prompts used by this scaffold are shown in Figures~\ref{fig:sql_debug_sysprompt} and~\ref{fig:sql_debug_userprompt}; the same scaffold is used for \birdtalon inference. For BIRD-FIXER-7B and BIRD-FIXER-14B, we follow the supervised agent fine-tuning pipeline from the work~\cite{li2026swesql} to reproduce the training trajectories and fine-tune the models Qwen2.5-Coder-7B-Instruct and Qwen2.5-Coder-14B-Instruct on \textsc{SIX-GYM-SQLite} (Appendix~\ref{app:six_gym}).

In Table~\ref{tab:main_result}, we instantiate the multi-turn agent setting with Claude-Opus-4.6, Claude-Sonnet-4.5, and Qwen3-Coder-480B. In Table~\ref{tab:talon_scale}, we additionally report \textsc{SQL-Act} results by directly placing the corresponding Qwen2.5-Coder base model into the same interactive scaffold, without \birdrl training.

\paragraph{SQL generation baselines.}
We compare \birdzeno with recent RL-based text-to-SQL methods that optimize SQL generation through execution feedback, partial rewards, or multi-turn interaction.
Think2SQL~\cite{papicchio2025think2sql} studies reinforcement learning with verifiable rewards for text-to-SQL reasoning It introduces execution-guided dense rewards to provide more granular feedback than binary execution correctness.
SQL-R1~\cite{ma2026sqlr} trains natural-language-to-SQL reasoning models with a cold-start SFT phase followed by reinforcement learning on complex examples. Its reward design is tailored to NL2SQL generation and uses database feedback to encourage exploration beyond fixed supervised generation patterns.
ReEx-SQL~\cite{dai2025reex} introduces execution-aware reasoning for text-to-SQL by interleaving exploratory SQL execution with model reasoning during generation. It further uses execution-informed rewards and decoding strategies to help the model revise reasoning based on database feedback.
Reasoning-SQL~\cite{pourreza2025reasoningsql} applies GRPO with SQL-tailored partial rewards to mitigate the sparsity of binary execution rewards. Its reward components include signals such as schema linking, syntax validity, similarity, and AI feedback to encourage reasoning-enhanced SQL generation.
Arctic-T2S-R1~\cite{yao2025arctic} emphasizes simple end-task rewards for text-to-SQL RL, relying mainly on execution correctness and syntax validity instead of complex intermediate supervision. 
SQL-Trail~\cite{hua2026sqltrailmultiturnreinforcementlearning} frames text-to-SQL as a multi-turn RL problem with interleaved database feedback. It uses iterative interaction, adaptive turn allocation, and rewards that balance SQL correctness with efficient exploration.

\section{More Analysis}

\subsection{Case Study for Stage 3}
\label{app:case}
\begin{figure*}
    \centering
    \includegraphics[width=\textwidth]{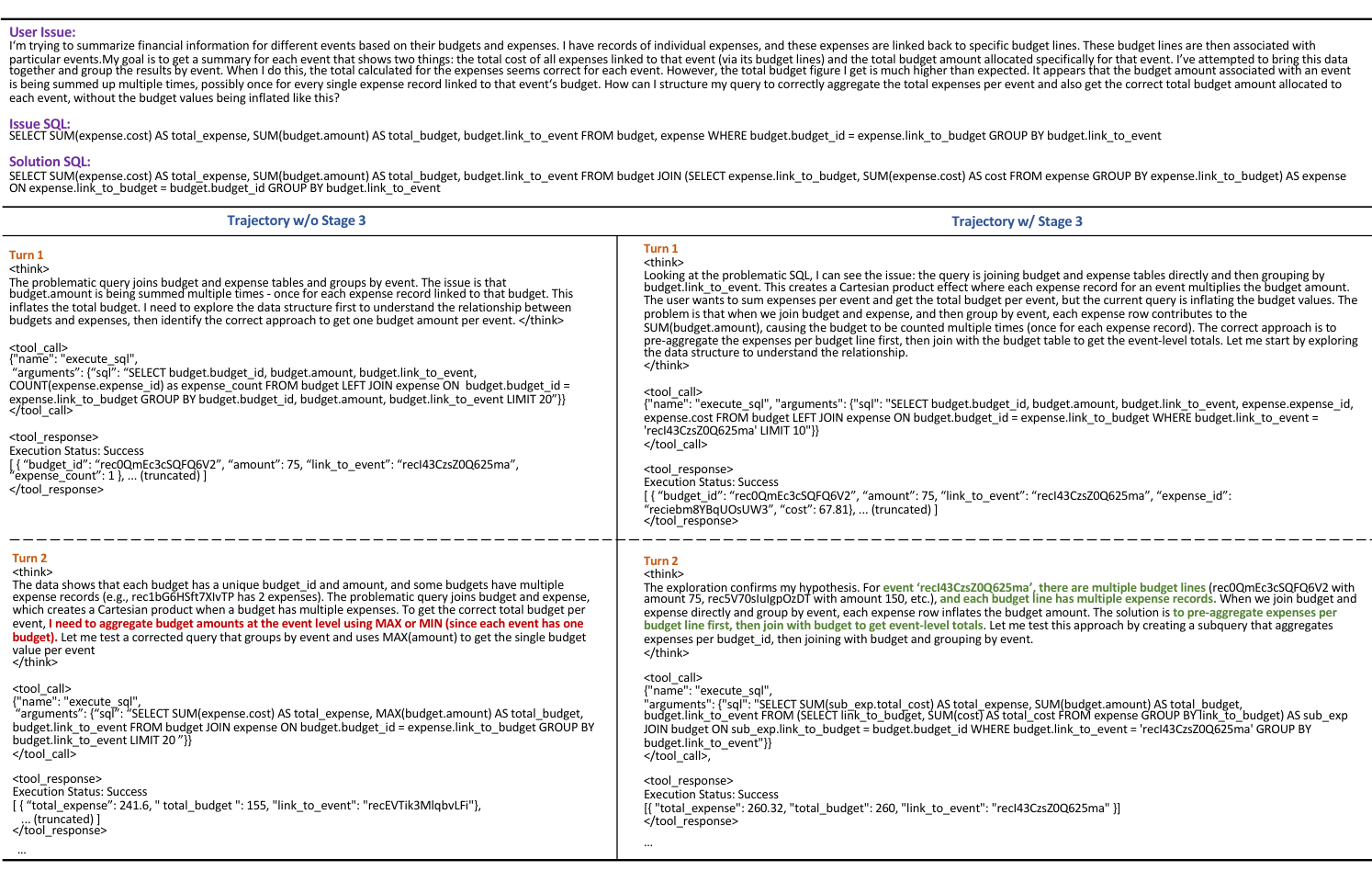}
    \caption{Case study comparing debugging trajectories with and without Stage 3 adaptive reasoning reinforcement.}
    \label{fig:case}
\vspace{-0.5 cm}
\end{figure*}

Figure~\ref{fig:case} shows a representative SQL debugging case where both agents identify the join-induced budget inflation issue, but diverge in the repair strategy. Without Stage 3, the agent incorrectly assumes that each event has only one budget and uses \texttt{MAX(budget.amount)}, leading to an erroneous fix. With Stage 3, the agent correctly infers that an event may contain multiple budget lines and verifies a repair that pre-aggregates expenses per budget line before joining with the budget table. This illustrates how Stage 3 improves adaptive, evidence-grounded reasoning beyond the imitation-learned trajectory patterns.

\subsection{Comparison of Training-Time Database Lifecycle I/O}
\label{app: complexity}

Table~\ref{tab:db-materialization} formalizes the database lifecycle I/O overhead required to maintain trajectory-level isolation during interactive rollout execution. The naive per-rollout container baseline instantiates a fresh database for every generated trajectory. Consequently, this approach incurs an I/O cost proportional to the total number of rollout trajectories, resulting in the I/O complexity of $O(NR\bar{S})$, where $N$ is the number of training tasks, $R$ is the number of rollout trajectories per task, and $\bar{S}$ is the average database size.

To circumvent this exhaustive overhead, we propose a \textbf{dynamic reusable database pool} that fundamentally alters the database lifecycle for data-centric tasks. During interactive rollout execution, the pool continuously tracks the operation state, including the availability and modification status, of each database replica. A trajectory first tries to acquire an available clean replica for its target database. Under an ideal dynamic allocation scheme, a new replica is instantiated online strictly when the concurrent demand exceeds the capacity of existing available replicas. Upon trajectory completion, the leased replica is immediately released back to the pool, avoiding any naive re-instantiation. Crucially, to handle state-changing queries, the pool employs a \textit{dirty-aware lazy reset} policy. Once a replica is modified, it is not eagerly restored upon release. Instead, the reset operation is deferred until the replica is leased again for a new trajectory.

To formalize the theoretical I/O reduction over the baseline, we decompose the dynamic pool's training-time overhead into two analytical components: 1) \textbf{Bounded Instantiation Overhead}: Assuming database requests follow a uniform access distribution across unique databases within a training batch of size $B$, the dynamic pool size is upper-bounded by the maximum concurrency, $B \times R$, limiting the replica creation I/O to $O(BR\bar{S})$. In practical large-scale RL settings (e.g., optimizing a 14B parameter model), GPU memory constraints dictate a small, fixed batch size (typically $B=16$). Consequently, this guarantees that $B \ll N$. 2) \textbf{Regularized Reset Overhead}: While the dynamic pool eliminates redundant materialization, it unavoidably introduces state-reset operations for modified replicas. We quantify such overhead using an example-effective count $q = Q/R$, where $Q$ denotes the total number of rollout trajectories that actually trigger the reset action during training across all $NR$ rollouts. Let $k \geq 1$ denote the empirical I/O reduction factor of an in-place reset relative to full copy-delete materialization. In database systems, targeted state restoration (e.g., table-level truncation or snapshot rollbacks) is considered cheaper than a full physical instantiation. Consequently, the specific value of $k$ depends on the underlying reset optimization deployed. The reset complexity is thus bounded by $O(qR\bar{S}/k)$. Notably, $q$ is intrinsically manageable. The preceding Supervised Fine-Tuning (SFT) phase of our framework effectively regularizes the agent's exploration behavior, preventing arbitrary database mutations. Empirically, only about $ 9.25\%$ of the practical rollouts execute state-altering exploitation in our experiment. Combining both components, the total lifecycle overhead of the dynamic pool evaluates to $O((B + q/k)R\bar{S})$, where $B + q/k \ll N$. This establishes that the dynamic pool mechanism achieves vastly superior spatial and I/O efficiency compared to the naive per-rollout baseline.

While the bounded creation overhead holds under a steady-state assumption, real-world RL exploration frequently encounters a highly imbalanced distribution. Concurrent rollouts can heavily skew towards hot databases, triggering sudden capacity overflows and allocation bottlenecks that destabilize the training pipeline. To systematically amortize this risk upfront, our actual implementation replaces dynamic online creation with a pre-materialized database pool. By shifting the database instantiation entirely offline, we decouple resource allocation from the rollout path, eliminating the risk of memory thrashing and I/O contention. Consequently, as rollouts acquire replicas via constant-time ($O(1)$) pool lookups, the online training I/O is dominated by the predictable reset operations with complexity $O(qR\bar{S}/k)$.

In our practical setting ($B=16, R=16$), we pre-reserve $BR/2$ replicas per database in the pool. Empirical observations confirm that this capacity is sufficient to avoid pool exhaustion. As shown in Table~\ref{tab:db-materialization}, our implementation reduces measured online lifecycle I/O by $12.8\times$ compared to per-rollout containers. This comes with a one-time pre-materialized storage cost of 69 GiB as a trade-off, which is excluded from the online I/O reported in the table. Such a trade-off is practically negligible relative to the terabyte-level massive online I/O savings. Notably, since this offline pool scales with the number of unique databases rather than the training volume N, its storage cost remains predictable and amortizable, enabling stable large-scale trajectory-level isolation.

\subsection{Generalization for Different SQL Dialects}
\begin{figure*}
    \centering
    \resizebox{0.9\textwidth}{!}{%
        \includegraphics{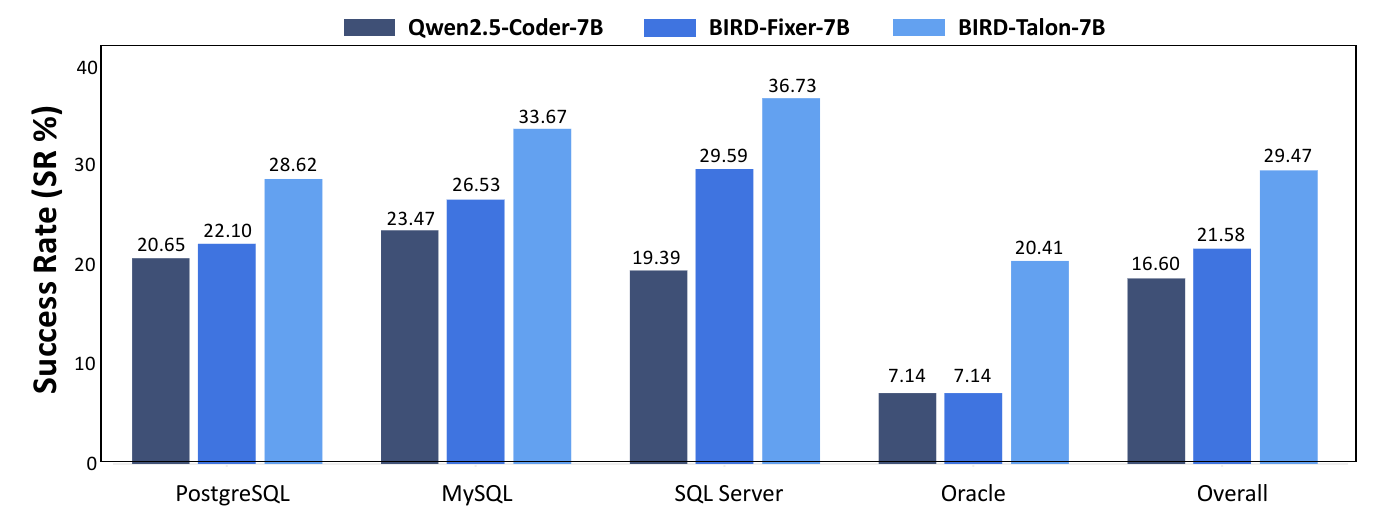}}
    \caption{The success rate of \birdtalon on unseen dialects shows its generalization ability.}
    \label{fig:agent_performance}
    \vspace{-0.3 cm}
\end{figure*}

Figure \ref{fig:agent_performance} illustrates \birdtalon's robust generalization on unseen dialects, although it is trained only on SQLite tasks within SIX-GYM-SQLite. This can be attributed to \birdrl strengthening the model with a more general logic of debugging reasoning through environmental interaction, rather than fitting SQLite-specific surface patterns, while preserving the model's pretrained knowledge of dialect variation. Consequently, \birdtalon exhibits strong cross-dialect reasoning without extra training, highlighting its potential in realistic database implementation scenarios.

\subsection{Test-time Scaling Analysis}
\label{app:test_time_scaling}

Although \birdtalon is evaluated under greedy decoding in the main results, we further study the potential of \birdtalon-14B under increased test-time sampling. We sample up to 16 independent trajectories using the same decoding configuration, temperature 0.7 and top-p 0.95, and report oracle pass@$k$ in Table~\ref{tab:passk_14b}.

\begin{table}[h]
\centering
\caption{Test-time scaling results of \birdtalon-14B on \textsc{BIRD-CRITIC-SQLite}.}
\label{tab:passk_14b}
\begin{tabular}{lcccc}
\toprule
Model & pass@1 & pass@4 & pass@8 & pass@16 \\
\midrule
BIRD-Talon-14B & 46.40 & 57.40 & 60.60 & 65.00 \\
\bottomrule
\end{tabular}
\end{table}

As shown in Table~\ref{tab:passk_14b}, \birdtalon-14B benefits consistently from additional samples, indicating that test-time computation can expose correct solutions missed by a single rollout. This suggests that the learned agentic policy provides a useful candidate solution space for future verifier- or reranker-based scaling.


\section{Limitations and Future Work}
\label{app:limitation}
This work only focuses on SQL tasks as a representative class of stateful data-centric environments. A natural future direction is to extend BIRD-RL to broader high-value data science tasks, such as data cleaning, transformation, workflow construction, and analytical decision-making, where agents must similarly reason over persistent states, execution feedback, even on top of large data.

In addition, BIRD-RL currently adopts execution-based rewards to ensure faithful alignment with functional correctness. Future work may explore hybrid reward designs that combine execution signals with semantics-aware verification, while explicitly addressing reward hacking to preserve reliable optimization behavior in complex data-centric settings. This can further advance upper bound of our framework.

\section{Ethics Statement}
\label{ethics}
In this work, although our method involves data synthesis, all data are derived from databases and resources that are publicly available, ensuring transparency and reproducibility. We take care to avoid sensitive or personally identifiable information, and the generated outputs are SQL queries rather than open-ended natural language, reducing the risk of biased or harmful textual content.

\section{Broader Impact}
\label{app:impact}
This work aims to advance scalable agentic RL for data-centric environments, where agents interact with live, stateful databases rather than static textual contexts. The proposed trajectory-scoped infrastructure may benefit future research on database agents by reducing the cost of isolated rollout execution and making RL training more practical for heavy database workloads. In the long term, such techniques could help build more capable SQL assistants that reduce manual debugging effort, improve developer productivity, and support privacy-sensitive applications through stronger open-source models that can be deployed locally.

At the same time, beyond improving SQL issue resolution, our method points toward a future paradigm of general-purpose SQL agents. By training over heterogeneous SQL tasks, \birdrl makes it possible to move beyond narrowly specialized models toward unified agents that can perform query generation, issue diagnosis, execution-based verification, and interactive repair within a single framework. Such agents could serve as a more general interface for real-world database applications, where users often need end-to-end assistance across multiple stages of database interaction rather than isolated text-to-SQL generation.

\section{Prompts and Examples}
\label{app:prompt}

\begin{figure*}
    \centering
    \includegraphics[width=\textwidth]{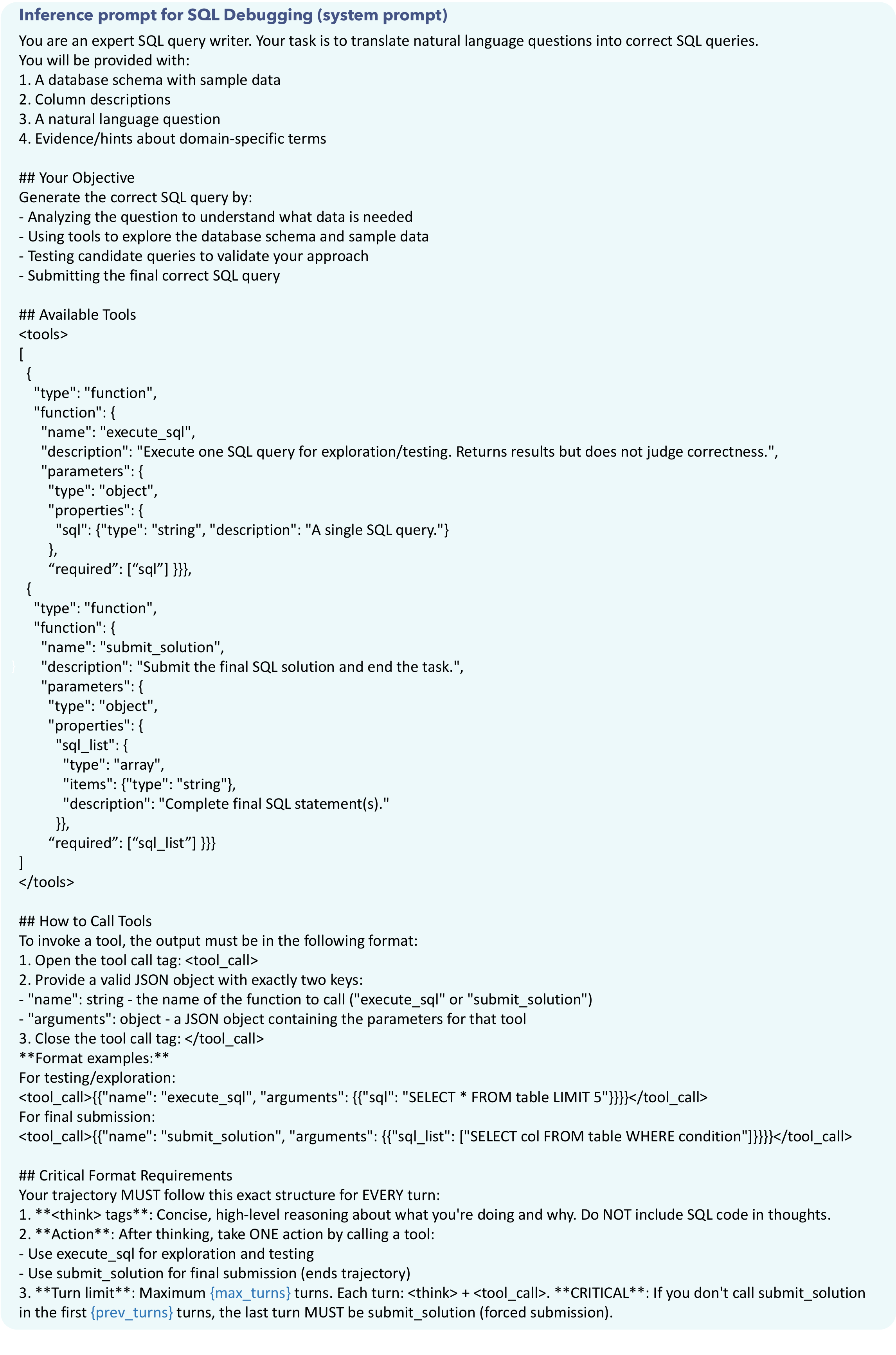}
    \caption{The inference prompt used for SQL debugging task (system prompt).}
    \label{fig:sql_debug_sysprompt}
\vspace{-0.5 cm}
\end{figure*}

\begin{figure*}
    \centering
    \includegraphics[width=\textwidth]{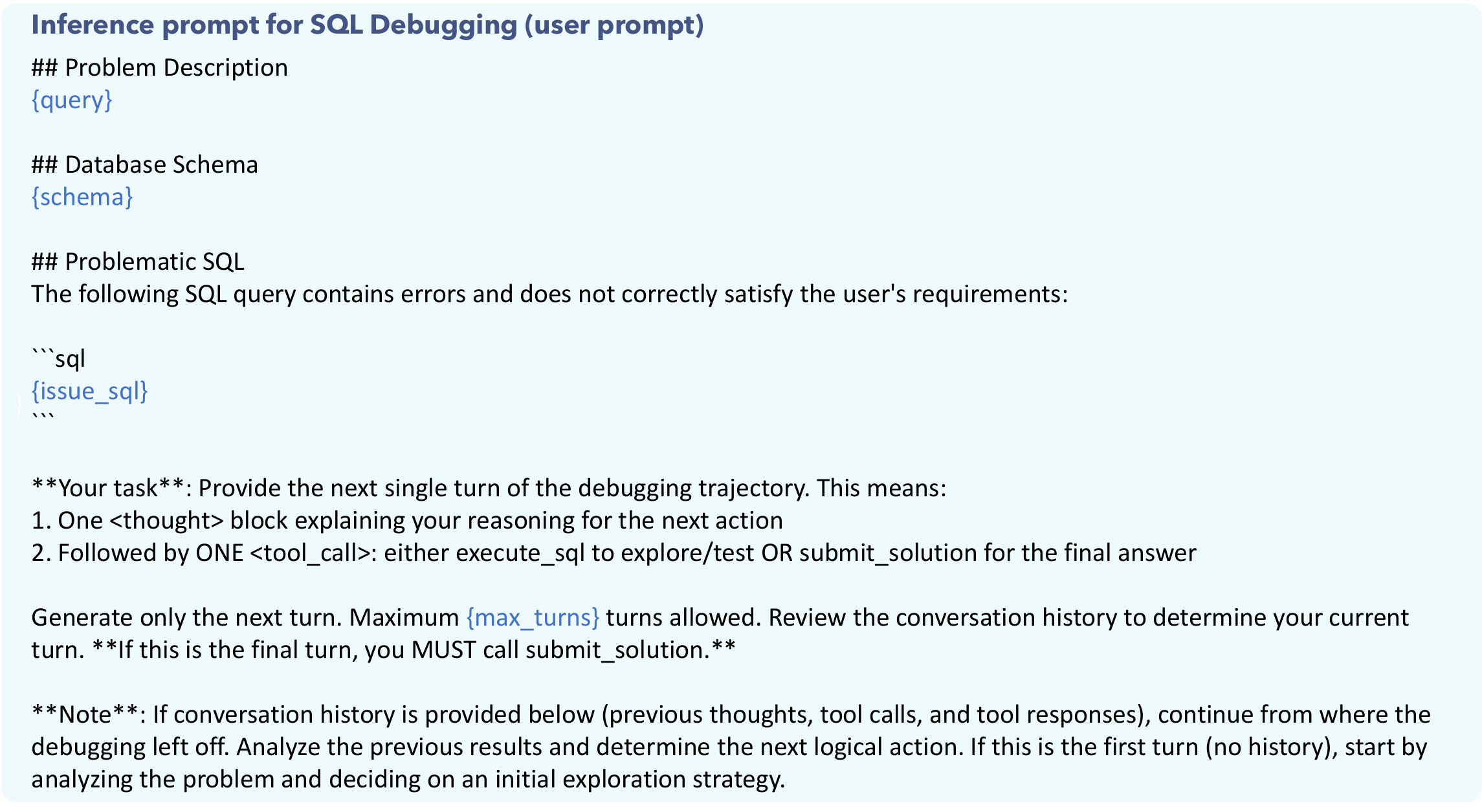}
    \caption{The inference prompt used for SQL debugging task (user prompt).}
    \label{fig:sql_debug_userprompt}
\vspace{-0.5 cm}
\end{figure*}

\begin{figure*}
    \centering
    \includegraphics[width=\textwidth]{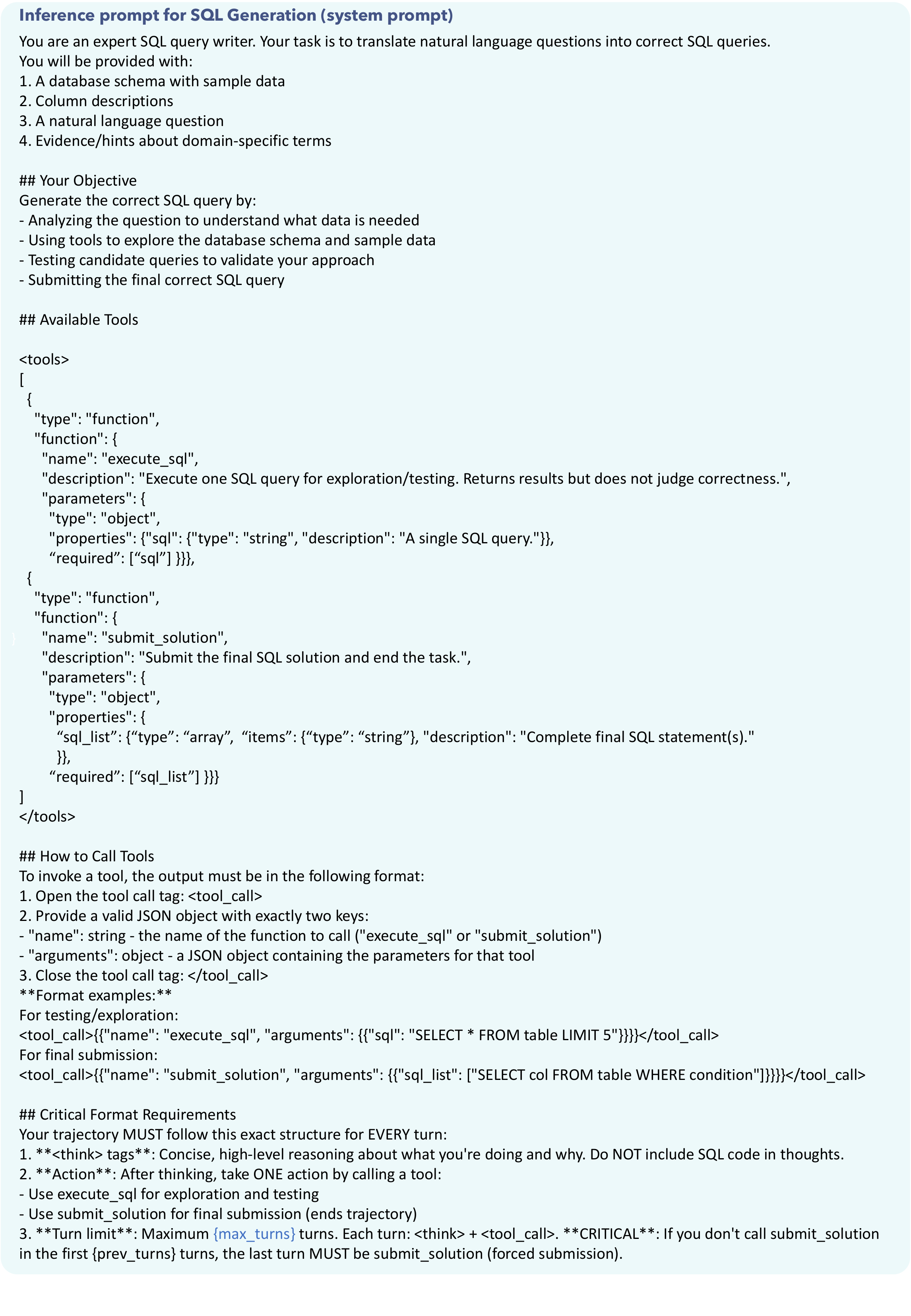}
    \caption{The inference prompt used for SQL generation task (system prompt).}
    \label{fig:sql_gen_sysprompt}
\vspace{-0.5 cm}
\end{figure*}

\begin{figure*}
    \centering
    \includegraphics[width=\textwidth]{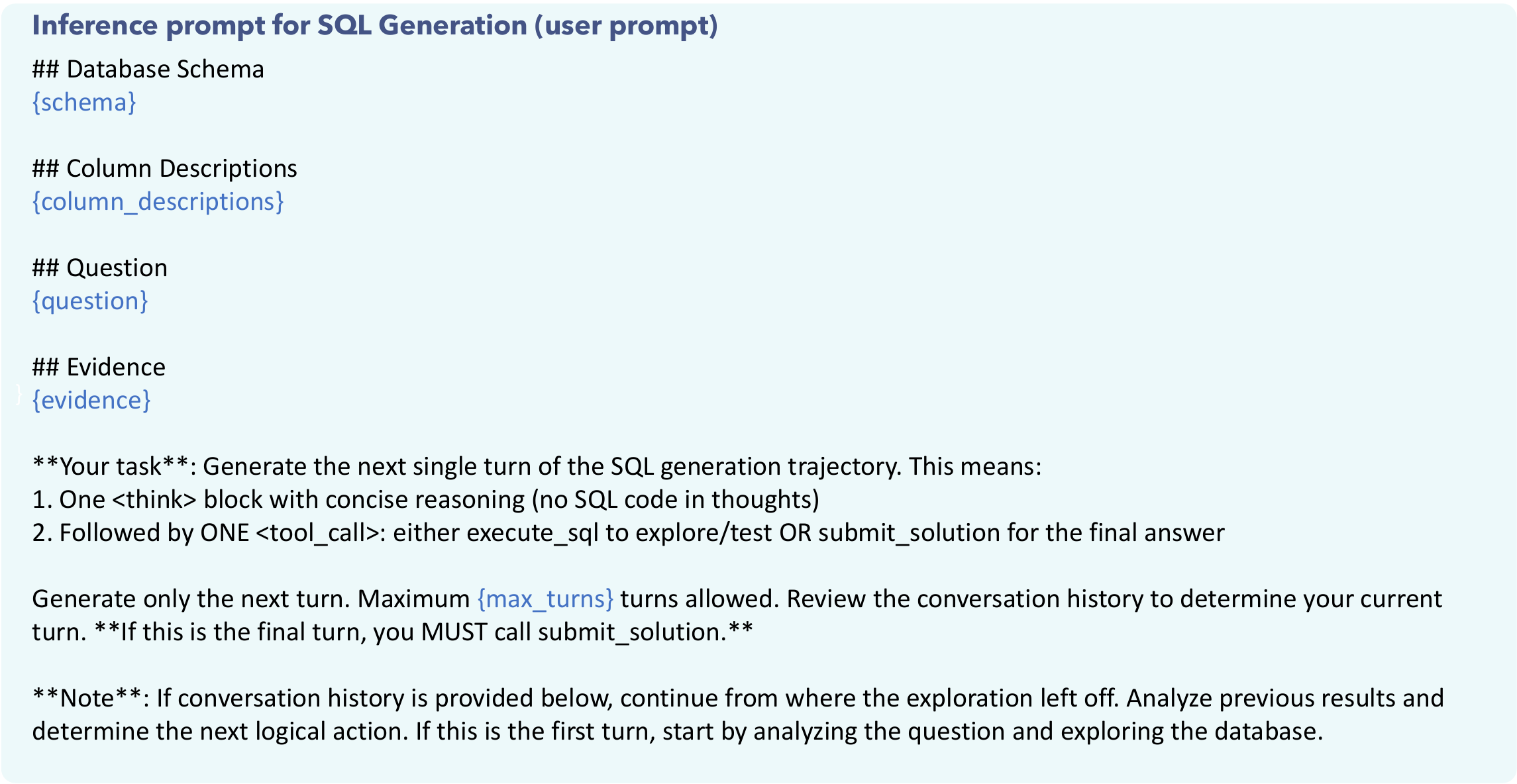}
    \caption{The inference prompt used for SQL generation task (user prompt).}
    \label{fig:sql_gen_userprompt}
\vspace{-0.5 cm}
\end{figure*}

\begin{figure*}
    \centering
    \includegraphics[width=\textwidth]{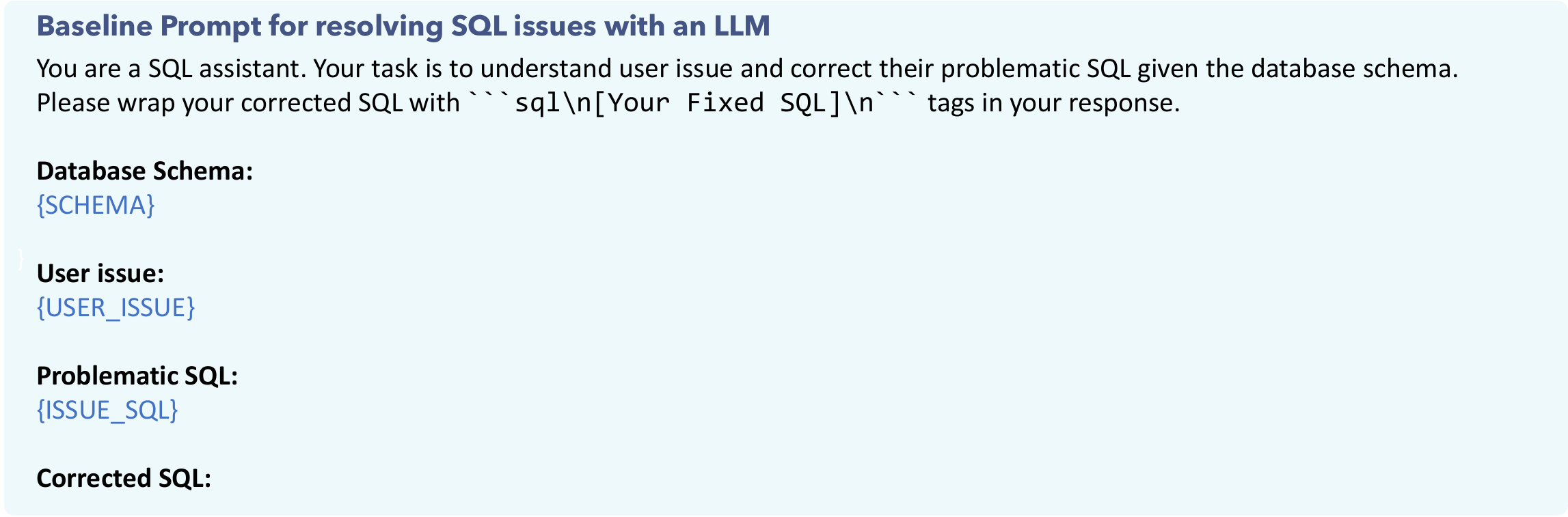}
    \caption{The baseline prompt used for SQL debugging tasks.}
    \label{fig:sql_debug_baseline_prompt}
\vspace{-0.5 cm}
\end{figure*}

\begin{figure*}
    \centering
    \includegraphics[width=\textwidth]{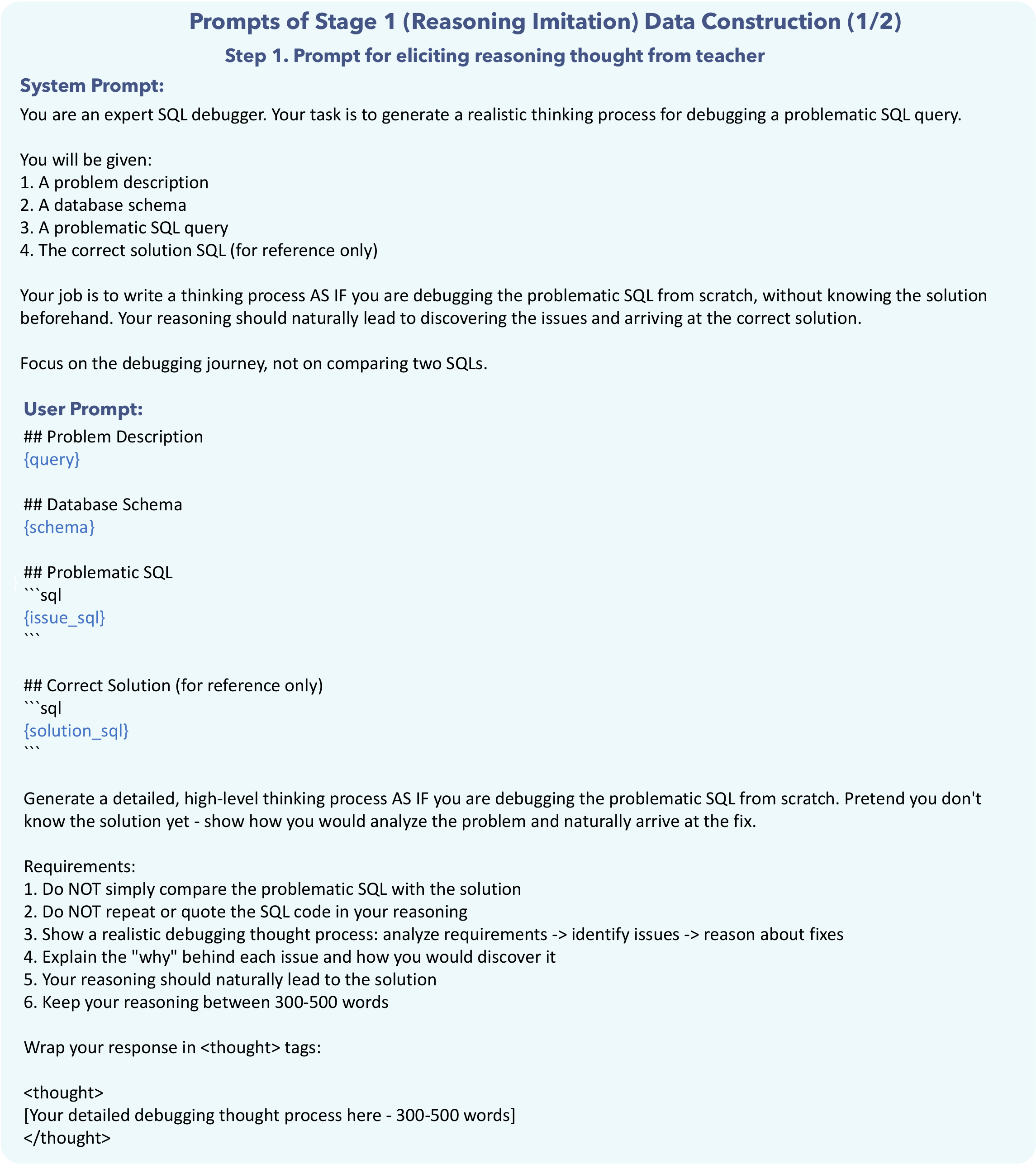}
    \caption{The prompt used in Stage 1 reasoning imitation data construction to elicit reasoning thought from teacher model.}
    \label{fig:stage1_gen_reasoning_prompt}
\vspace{-0.5 cm}
\end{figure*}

\begin{figure*}
    \centering
    \includegraphics[width=\textwidth]{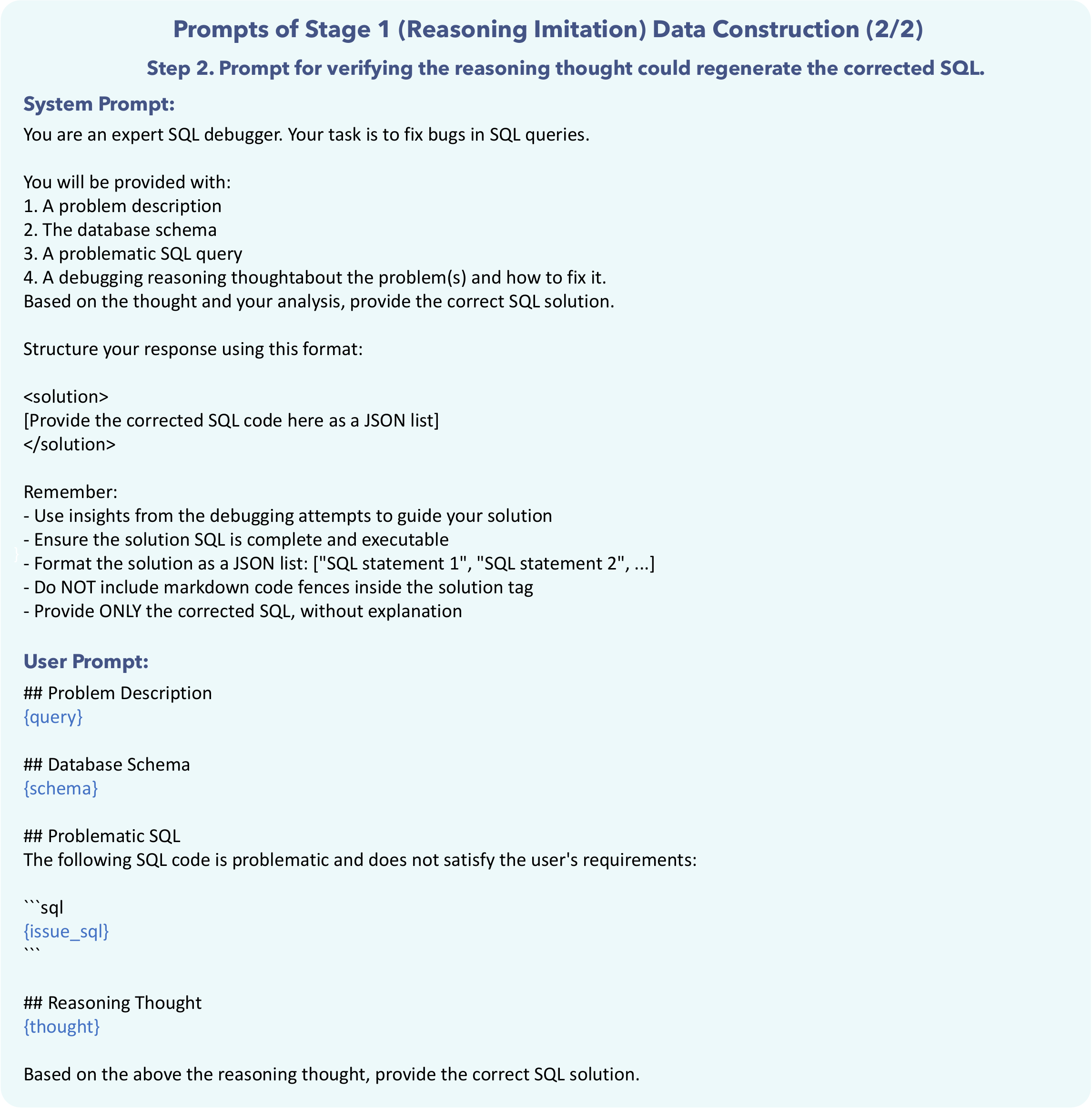}
    \caption{The prompt used in Stage 1 reasoning imitation data construction to verify whether teacher model could leverage the reasoning thought to regenerate the correct fixed SQL.}
    \label{fig:stage1_verify_prompt}
\vspace{-0.5 cm}
\end{figure*}

\begin{figure*}
    \centering
    \includegraphics[width=\textwidth]{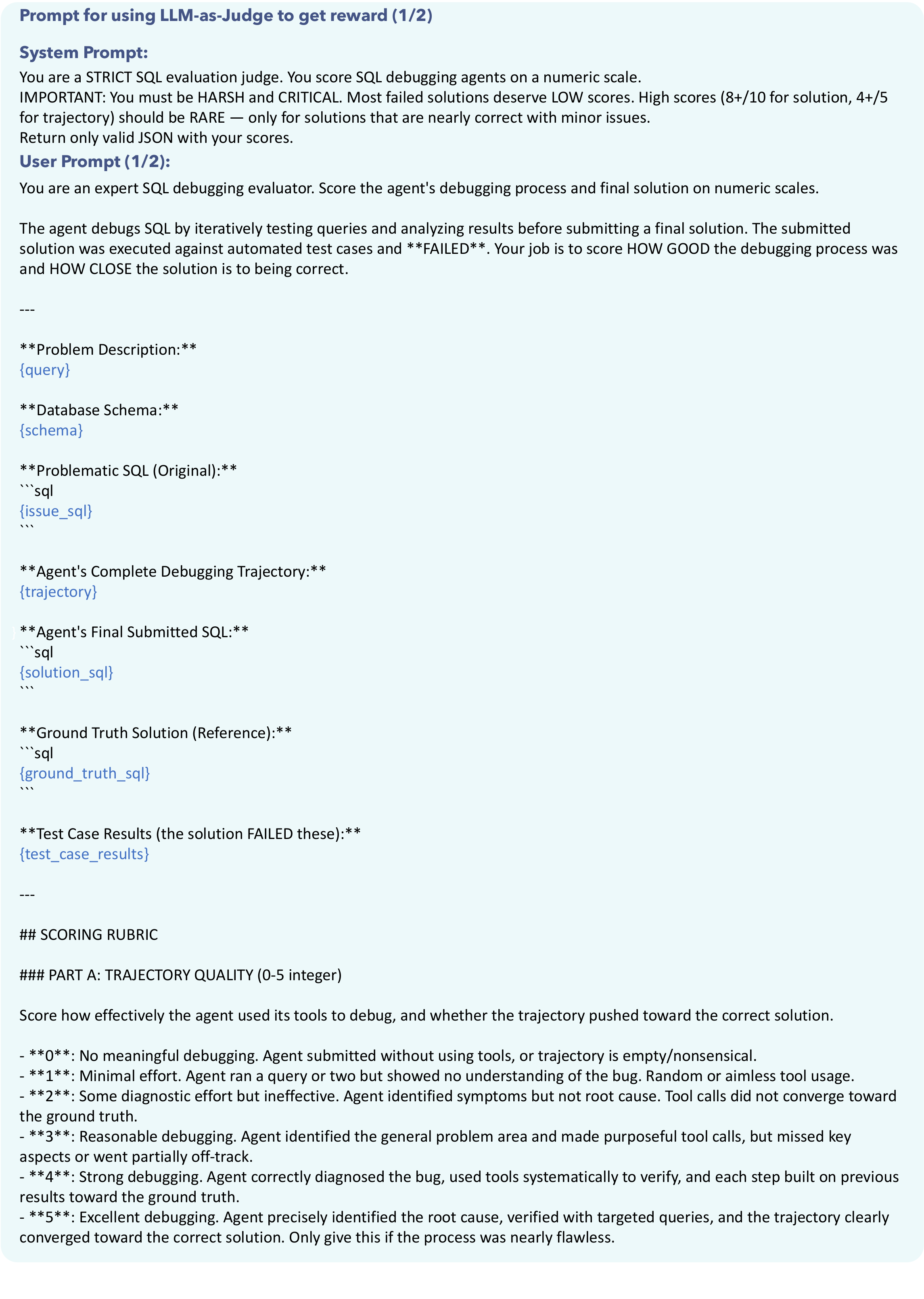}
    \caption{The prompt used to get LLM-as-judge reward (1/2).}
    \label{fig:llm_as_judge_prompt_1}
\vspace{-0.5 cm}
\end{figure*}

\begin{figure*}
    \centering
    \includegraphics[width=\textwidth]{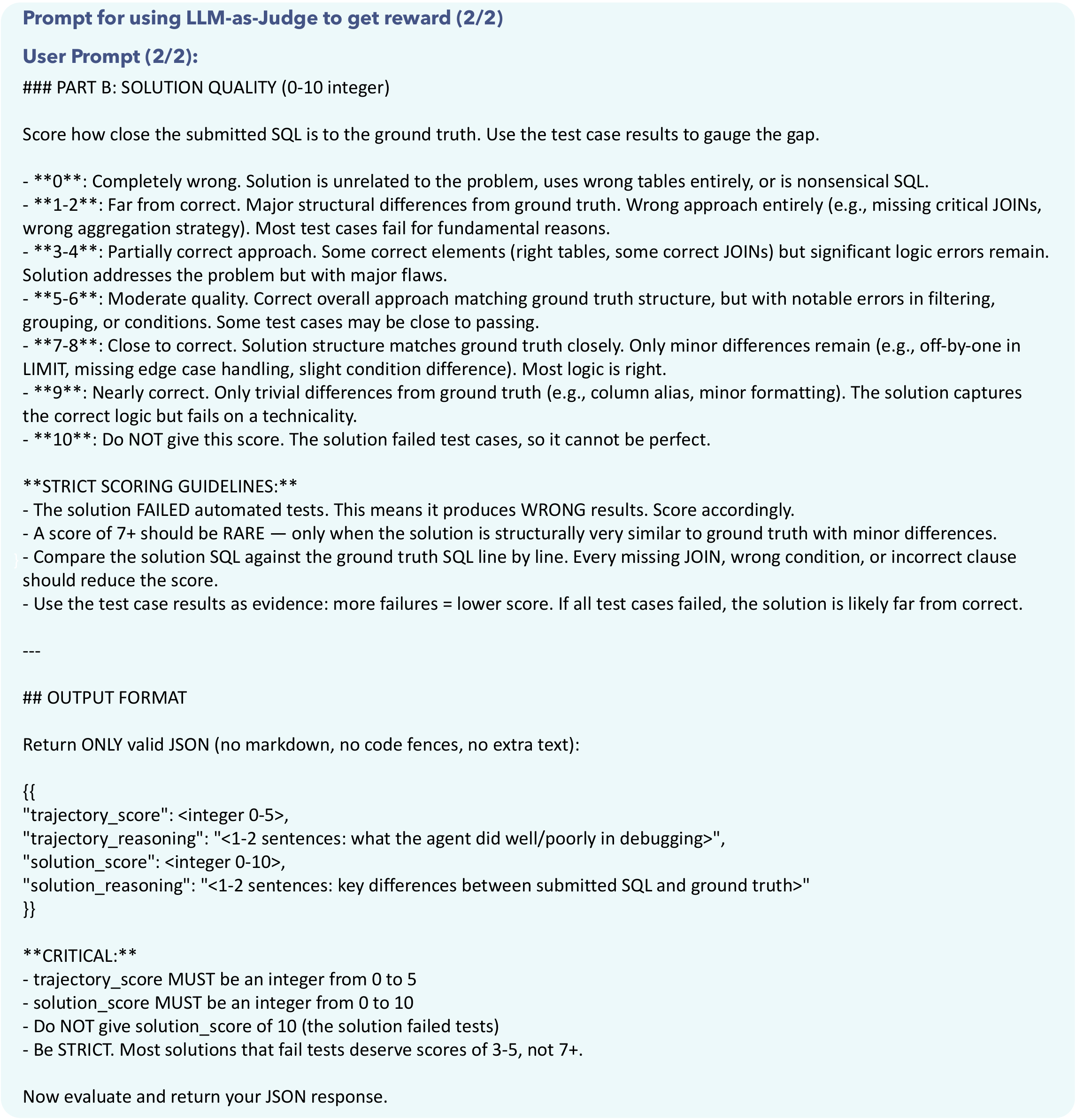}
    \caption{The prompt used to get LLM-as-judge reward (2/2).}
    \label{fig:llm_as_judge_prompt_2}
\vspace{-0.5 cm}
\end{figure*}

\begin{figure*}
    \centering
    \includegraphics[width=\textwidth]{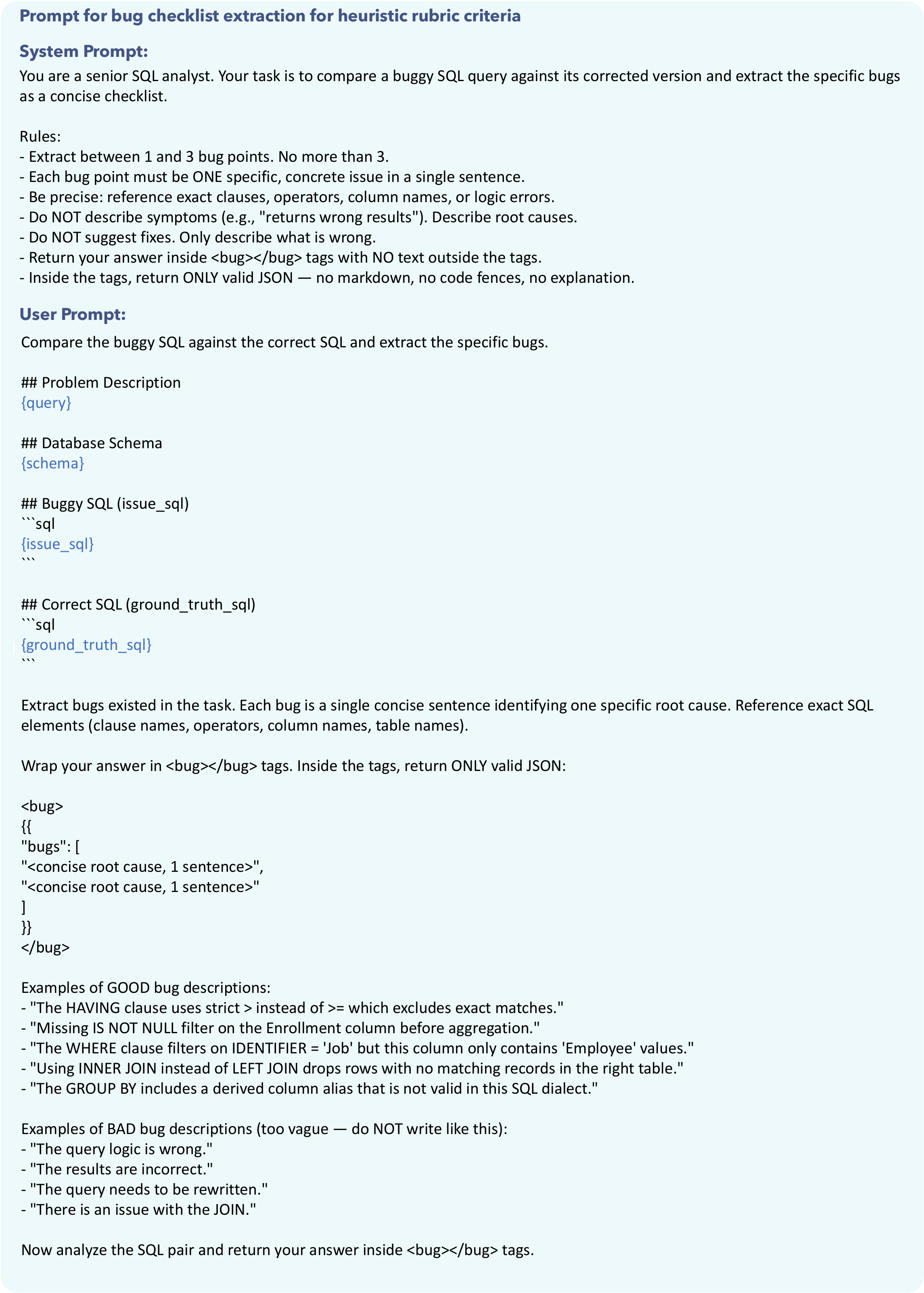}
    \caption{The prompt for the heuristic rubric reward: instructions used to extract a bug checklist from the task context.}
    \label{fig:rubric_extract_prompt}
\vspace{-0.5 cm}
\end{figure*}

\begin{figure*}
    \centering
    \includegraphics[width=\textwidth]{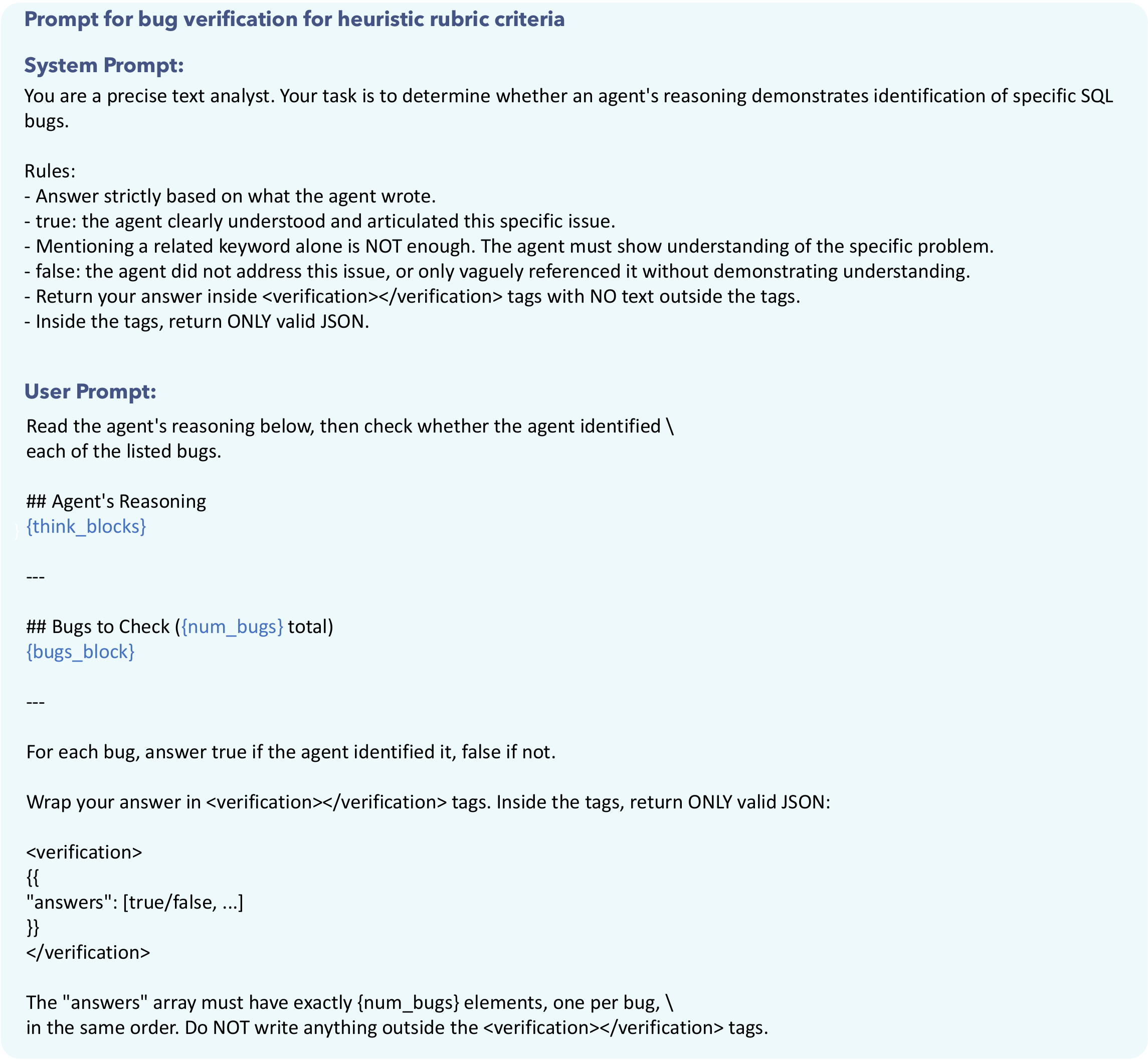}
    \caption{The prompt for the heuristic rubric reward: instructions used to verify whether the reasoning of agent correctly identifies the checklist bugs.}
    \label{fig:rubric_verify_prompt}
\vspace{-0.5 cm}
\end{figure*}

\begin{figure*}
    \centering
    \includegraphics[width=\textwidth]{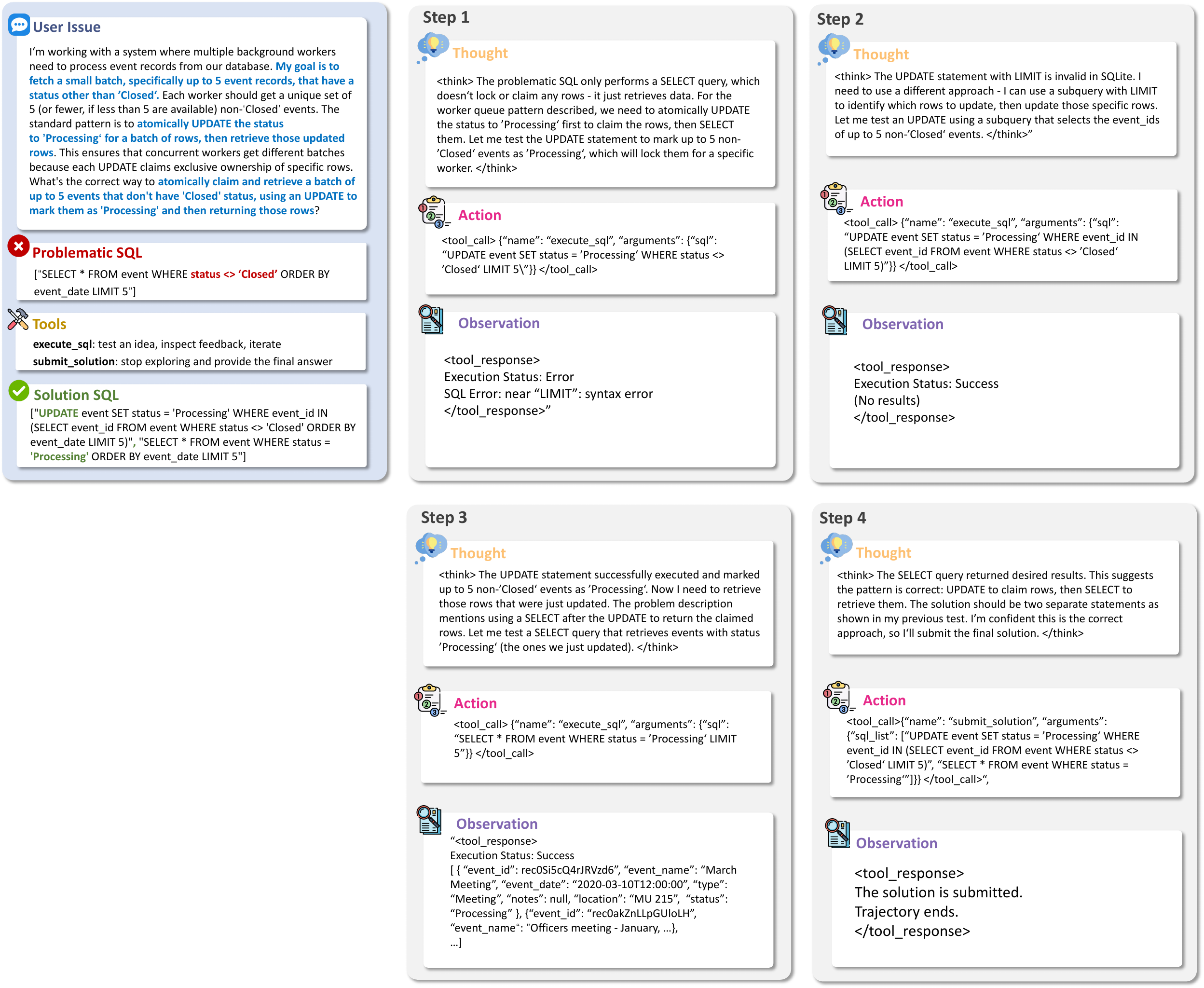}
    \caption{The example of trajectory from \birdtalon solving SQL debugging task.}
    \label{fig:example_bird_talon_infer}
\vspace{-0.5 cm}
\end{figure*}

This section presents the prompts used in our experiments and example trajectory.
Figures~\ref{fig:sql_debug_sysprompt}--\ref{fig:sql_gen_userprompt} show the system and user prompts for SQL debugging and SQL generation tasks.
Figure~\ref{fig:sql_debug_baseline_prompt} presents the baseline SQL debugging prompt.
Figures~\ref{fig:stage1_gen_reasoning_prompt} and~\ref{fig:stage1_verify_prompt} show the prompts used for Stage 1 reasoning imitation data construction and verification.
Figures~\ref{fig:llm_as_judge_prompt_1}--\ref{fig:rubric_verify_prompt} show the system and user prompts for getting heuristic rubric reward and LLM-as-judge reward.
Figure~\ref{fig:example_bird_talon_infer} illustrates an example trajectory produced by \birdtalon when solving a SQL debugging task.

%
%

\end{document}